\documentclass{article}

\usepackage[preprint]{neurips_2025}

\usepackage[utf8]{inputenc} 
\usepackage[T1]{fontenc}    
\usepackage{hyperref}       
\usepackage{url}            
\usepackage{booktabs}       
\usepackage{amsfonts}       
\usepackage{amsmath}        
\usepackage{nicefrac}       
\usepackage{microtype}      
\usepackage{graphicx}
\usepackage{amssymb}
\usepackage{algorithm}
\usepackage{algpseudocode}
\usepackage{bbm}
\usepackage{times}
\usepackage{latexsym}
\usepackage{multirow}
\usepackage{booktabs}
\usepackage{diagbox}
\usepackage{enumitem}
\usepackage{subcaption}
\usepackage{bm}
\usepackage{natbib}
\setcitestyle{numbers,square}
\PassOptionsToPackage{numbers, compress}{natbib}
\usepackage[table]{xcolor}
\graphicspath{ {./figures} }

\DeclareMathOperator*{\argmin}{arg\,min}

\newcommand{\model}[1]{FLoE}

\title{\model~: Fisher-Based Layer Selection for Efficient Sparse Adaptation of Low-Rank Experts}

\author{%
  Xinyi Wang \\
  Zhejiang University \\
  \texttt{wxyzju@zju.edu.cn} \\
  \And
  Lirong Gao \\
  Zhejiang University \\
  \texttt{gaolirong@zju.edu.cn}
  \And
  Haobo Wang \\
  Zhejiang University \\
  \texttt{wanghaobo@zju.edu.cn} \\
  \AND
  Yiming Zhang \\
  Zhejiang University \\
  \texttt{yimingz@zju.edu.cn} \\
  \And
  Junbo Zhao \\
  Zhejiang University \\
  \texttt{j.zhao@zju.edu.cn} \\
}

\begin{document}

\maketitle

\begin{abstract}
Parameter-Efficient Fine-Tuning (PEFT) methods have emerged as a widely adopted strategy for adapting pre-trained Large Language Models (LLMs) to downstream tasks, significantly reducing memory and computational costs. However, most existing PEFT techniques uniformly deploy LoRA adapters across all layers, disregarding the intrinsic heterogeneity of layer contributions and task-specific rank requirements. This uniform paradigm leads to redundant parameter allocation and suboptimal adaptation efficiency. To address these limitations, we propose \model~, a novel PEFT framework that introduces two key innovations: (i) a Fisher information-guided importance scoring mechanism to dynamically identify task-critical transformer layers for MoE-based low-rank adaptation, enabling sparse adapter deployment; and (ii) a Bayesian optimization-driven rank allocator that automatically determines optimal LoRA ranks on specific datasets without exhaustive grid search. Extensive experiments across diverse LLMs and benchmarks reveal that \model~ achieves impressive efficiency-accuracy trade-offs, making \model~ particularly advantageous in resource-constrained environments that necessitate rapid adaptation.
\end{abstract}

\section{Introduction}
Adapting Large Language Models (LLMs) for multiple downstream tasks traditionally relies on full fine-tuning (FFT), which requires retraining all model parameters.
To reduce the training cost, parameter-efficient fine-tuning (PEFT) techniques~\citep{meng2404pissa,hayou2024lora,ding2023sparse} have been developed, which can be broadly categorized into LoRA-based~\citep{hayou2024lora,ding2023sparse,liu2024dora}, Adapter-based~\citep{wang2022adamix,lei2023conditional} and Prompt-based~\citep{li2021prefix,liu2021p,lester2021power} approaches.
While tuning a limited set of parameters is effective for domain adaptation, PEFT methods like LoRA~\citep{hu2021lora} often exhibit a performance gap compared to the FFT baseline.
This gap widens further when tuning on complex datasets~\citep{longpre2023flan} with diverse sub-domains and task types, which requires models to distinguish subtle, non-overlapping features while avoiding redundancy.

\begin{figure}[ht]
    \centering
    \begin{subfigure}[t]{0.48\linewidth}
        \includegraphics[width=\linewidth]{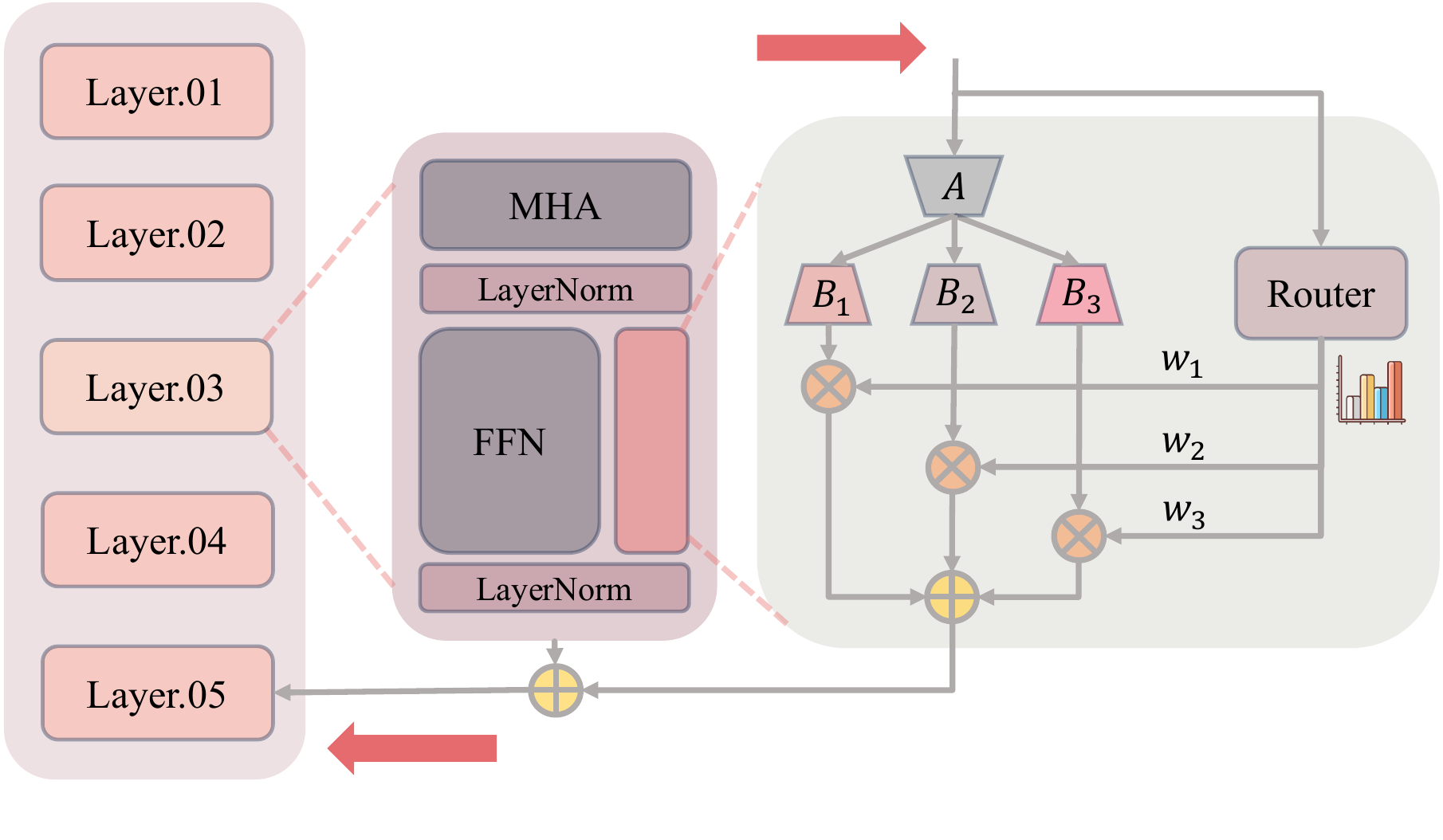}
        \vspace{-0.5em}
        \caption{}
        \label{fig:adapter_struc}
    \end{subfigure}
    \hfill
    \begin{subfigure}[t]{0.48\linewidth}
        \includegraphics[width=\linewidth]{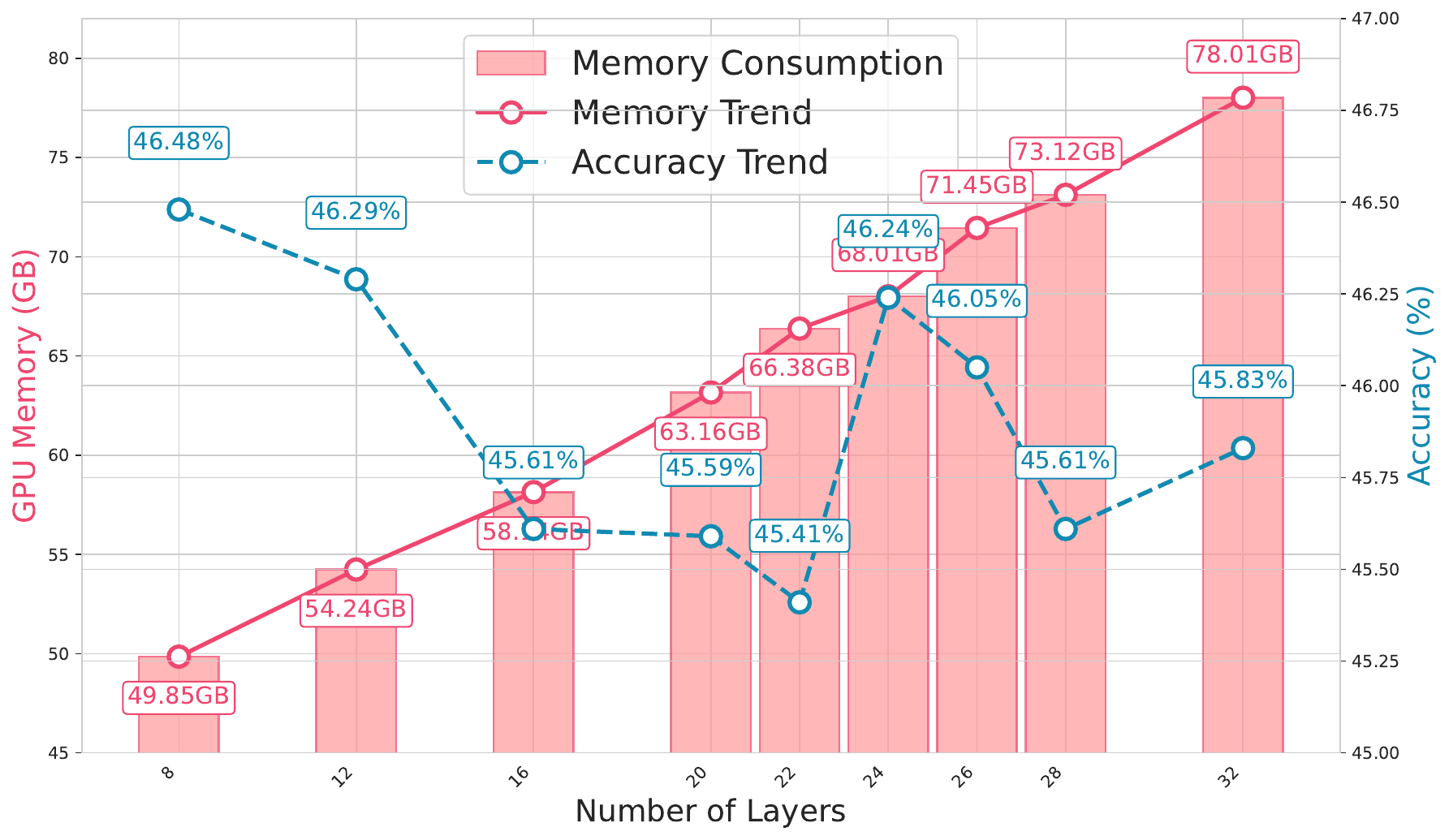}
        \vspace{-0.5em}
        \caption{}
        \label{fig:gpu_sub}
    \end{subfigure}
    \caption{(a) Architecture of our MoE-based LoRA implementation. We adopt the asymmetric architecture of HydraLoRA~\citep{tian2024hydralora}: a shared $A$ matrix captures general features of the dataset, and multiple distinct low-rank experts $B_i$ learn task-specific patterns. The router takes in an intermediate token representation and generates gating scores $w_i$ over experts.
    (b) Experimental evidence supporting our motivation: we evaluate GPU memory consumption and adaptation performance of the \model~ layer selection algorithm on LLaMA2-7B fine-tuned with Databricks-Dolly-15K. As the number of adapted layers increases, GPU memory usage grows linearly, yet model accuracy does not improve and even underperforms compared to few-layer adaptation.}
    \label{fig:gpu_and_norm}
\end{figure}

Recent studies explore a hybrid solution~\citep{gao-etal-2025-mola,dou2023loramoe,huang2023lorahub}, showing that combining LoRA with the Mixture-of-Experts (MoE)~\citep{jacobs1991adaptive,shazeer2017outrageously} is a promising recipe.
Among these solutions, HydraLoRA~\citep{tian2024hydralora} stands out by discovering the asymmetric property of LoRA and implementing $B$ matrices as domain-specific experts, achieving impressive adaptation performance.
However, existing methods~\citep{tian2024hydralora,zhang2023adalora,liu2024dora,dou2023loramoe} adopt a uniform placement strategy that indiscriminately deploys fixed-rank LoRA adapters across all transformer layers.
Our investigation yields two critical observations that challenge the premise of current implementations:
\begin{itemize}[topsep=0pt,parsep=0pt,partopsep=0pt,leftmargin=2em]
    \item Indiscriminate deployment of MoE-based LoRA adapters leads to unnecessary computational overhead as shown in Fig.~\ref{fig:gpu_sub}, revealing a paradoxical trade-off between the number of trainable parameters and overall performance gains.
    \item LoRA-based tuning is highly sensitive to the choice of rank~\citep{zhang2023adalora,jiang2024mora,valipour2022dylora,liu2024alora}. Since models trained with one rank do not generalize to others, it is crucial to identify the optimal rank in advance to avoid costly retraining for each possible rank.
\end{itemize}

Based on these observations, the key to improve MoE-based LoRA tuning is to identify and adapt a small number of critical layers. To quantify the concept of "critical", our motivation is that: if a layer is important for task-specific adaptation, the parameters of its residual trainable adapter should exhibit high sensitivity to the adaptation loss, while the corresponding pre-trained weights remain relatively insensitive.
From a mathematical view, we can use the variation of gradients to quantify this property, which can be measured with Fisher Information.
Another question is how to determine the optimal rank before training.
To address these questions, we propose \model~, a sparse layer adaptation framework that provides a unified selection of layer and LoRA rank.
The overall \textit{pipeline} includes: first fine-tune a full-layer model on a sampled dataset using MoE-based LoRA shown in Figure~\ref{fig:adapter_struc}, then applies \model~ to determine critical layers and optimal ranks. During final adaptation on the target dataset, all pre-trained weights are frozen while only the adapters on the critical layers are updated.

Our contributions can be summarized as follow:
\begin{itemize}[topsep=0pt,parsep=0pt,partopsep=0pt,leftmargin=2em]
\item We introduce a Fisher-based importance scoring algorithm that dynamically identifies critical transformer layers for MoE-based low-rank adaptation, enabling sparse, context-aware adapter deployment.
\item We incorporate a Bayesian optimization step to estimate the optimal LoRA rank before training on the target dataset, avoiding exhaustive grid search and retraining.
\item Experiments show that \model~ achieves comparable or even better performance than prior PEFT methods across diverse datasets and model families, with notable advantages in low-resource and fast-adaptation scenarios. By adapting only 25\% of layers, \model~ retains 93.1\% of full fine-tuning accuracy on MMLU benchmarks, and achieves a 7.0\% relative improvement over the best-performing full-layer methods in mixed-domain adaptation, demonstrating its superior capability in mitigating domain interference while maintaining parameter efficiency.
\end{itemize}

\section{Related Work}
\noindent{\textbf{Parameter-Efficient Fine-tuning.}}
Parameter Efficient Fine-Tuning (PEFT) techniques aim to reduce the training costs of the LLMs.
Previous PEFT approaches can be broadly classified into the following categories:
i) Prefix-tuning~\citep{li2021prefix} and prompt-tuning~\citep{lester2021power}: prominent approaches that fine-tune continuous prompts rather than discrete ones.
ii) Adapter-based tuning: inserts additional adapters into the model or scales activations with learned vectors, including (IA)$^3$~\citep{liu2022few} for few-shot settings and AdaMix~\citep{wang2022adamix} for mixture of adaptations.
iii) Low-rank adaptation: introduces trainable low-rank matrices to LLMs, keeping the original weights frozen for efficiency, including LoRA~\citep{hu2021lora} and its variants, such as AdaLoRA~\citep{zhang2023adalora}, HydraLoRA~\citep{tian2024hydralora} and others~\citep{zhang2023lora,kopiczko2023vera,renduchintala2023tied,chen2023longlora,liu2024dora}.
Extensions to multi-LoRA architectures include Multi-Head Routing~\citep{page2024multi} for Mixture-of-Experts and LoraHub~\citep{huang2023lorahub} for task composability.

\noindent{\textbf{Layer-wise Selective Fine-tuning.}}
Recent studies~\citep{elhoushi2024layer,sajjad2023effect,zhang2023crash} have raised the issue of layer redundancy in pre-trained models. Surgical fine-tuning~\citep{lee2022surgical} updates only a subset of layers based on domain shift, while SubTuning~\citep{kaplun2023less} employs a greedy search to identify the most suitable layers, requiring significant computational resources.
For layer-wise sparse training, LISA~\citep{pan2024lisa} dynamically optimizes important layers based on weight norm, achieving faster convergence and improved performance.

\section{Methodology}
\begin{figure}[tb!]
    \centering
    \includegraphics[width=0.68\linewidth]{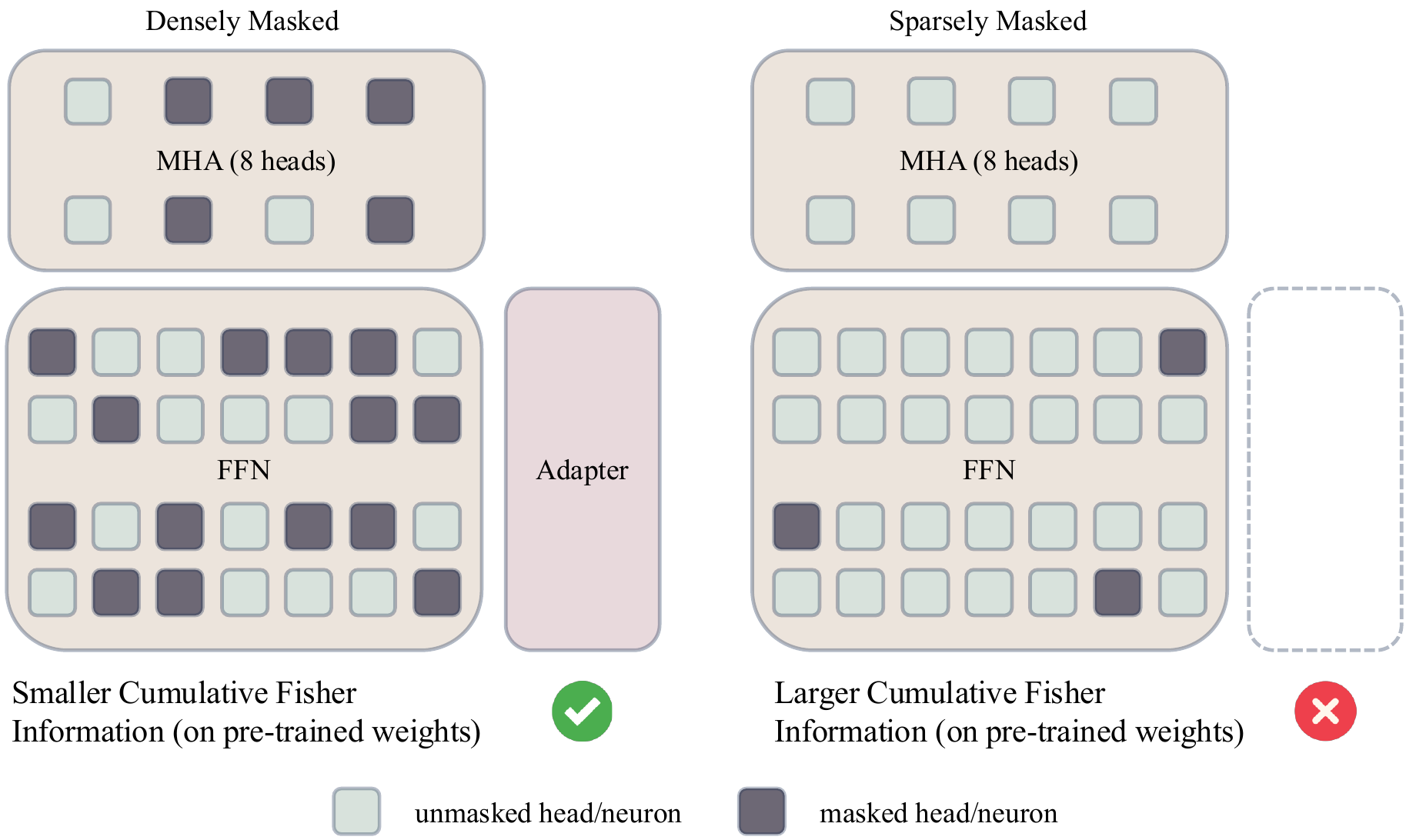}
    \caption{Mechanism of \model~ layer selection. A densely masked layer intends to have higher importance for low-rank adaptation, so we add a residual trainable adapter to the FFN component. The adaptation process only updates the adapter.}
    \label{fig:mask_estimation}
\end{figure}

Let $\mathcal{F}$ denote a pre-trained $L$-layer Transformer model. Given a dataset $\mathcal{D}=\{(x_i,y_i)\}_{i=1}^N$, where $x_i$ denotes the input data and $y_i$ the corresponding label. For layer $k \in \{1, \ldots, L\}$, let $\theta_k$ denote its pre-trained weights. LoRA introduces trainable low-rank matrices $A_k \in \mathbb{R}^{d \times r}$ and $B_k \in \mathbb{R}^{r \times d}$ to approximate weight updates $\phi_k=B_kA_k$.
Here we follow HydraLoRA~\citep{tian2024hydralora} to use an MoE-based architecture, extending LoRA by employing a shared $A$ matrix and $M$ parallel low-rank experts $\{B_k^{(i)}\}_{i=1}^M$ alongside a router network (implemented as a dense layer followed by a softmax function). Suppose the router outputs a vector of contribution weights $\{\omega_k^{(i)}\}_{i=1}^M$ based on the intermediate token representation. The weight updates are formulated as:
\begin{equation}
    \phi_k = \sum_{i=1}^M\omega_k^{(i)}B_k^{(i)}A_k
\end{equation}
The final merged weights are $W_k = \theta_k + \phi_k$.
\subsection{Problem Formulation}
As our goal is to find a subset of layers $\mathcal{S}\subseteq\{1,2,\ldots,L\}$ to add trainable adapters, we quantify the contribution of each layer to the model adaptation performance via an importance score $s_k$.
To achieve this, we introduce a binary mask variable $\mathbf{m}_k \in \{0,1\}^{|\theta_k|}$ as an intermediate to calculate $s_k$ during adaptation.
The adapted weight $\tilde{\theta}_k$ is then computed as:
\begin{equation}
    \tilde{\theta}_k = \mathbf{m}_k \odot \theta_k + \phi_k,
\end{equation}
As shown in Figure~\ref{fig:mask_estimation}, the mask variable $\mathbf{m}_k$ is applied to the pre-trained weights $\theta_k$, based on the principle that mask sparsity reflects the necessity of adaptation. If a pre-trained weight $\theta_{k,i}$ is masked ($\mathbf{m}_{k,i}=0$), the adapter weights must compensate for its removal to maintain performance. If $\theta_{k,i}$ is retained ($\mathbf{m}_{k,i}=1$), then $\phi_{k,i}$ only serves as a residual correction.
Thus, the sparsity of $\mathbf{m}_k$ directly correlates with the contribution of $\phi_k$. Let $\|\mathbf{m}_k\|_0$ denote the number of masked weights.
Higher sparsity ($\|\mathbf{m}_k\|_1 \ll \|\mathbf{m}_k\|_0$) indicates that $\theta_k$ are poorly aligned with the dataset, requiring significant adapter intervention.
Lower sparsity ($\|\mathbf{m}_k\|_0 \ll \|\mathbf{m}_k\|_1$) suggests the pre-trained knowledge in layer $k$ remains largely valid, requiring minimal adaptation. This explicitly disentangles pre-trained knowledge retention from task-driven adaptation.

Let $\bm{\theta} \triangleq [\theta_1,\ldots,\theta_L]$ and $\bm{\phi} \triangleq [\phi_1,\ldots,\phi_L]$.
The fine-tuning process only updates $\bm{\phi}$ while keeping $\bm{\theta}$ frozen. After obtaining $\bm{\theta}$ and $\bm{\phi}$, we optimize the mask variable $\mathbf{m}$ to find the optimal $\mathcal{S}$ under the following constrained objective:
\begin{equation}\label{eq:obj}
    \underset{\mathbf{m}}{\argmin}\ \mathcal{L}(\mathbf{m}; \bm{\theta} + \bm{\phi})\quad\text{s.t.}\quad\text{Cost}(\mathbf{m}; \bm{\theta}) \leq C
\end{equation}

\subsection{Fisher Information-Aware Estimation}
To enable gradient-based optimization for the constrained problem Eq.~\ref{eq:obj}, the cost function should be differentiable with respect to the mask $\mathbf{m}$.
Here, we use Taylor importance~\citep{ma2023llm} as the cost function, which measures the sensitivity of pre-training loss to parameter perturbations. This allows us to identify parameters that have minimal influence on the base model prediction, as indicated by the deviation in the pre-training loss.
For layer $k$, its element-wise Taylor Importance $T_k$ is defined as:
\begin{equation}
    T_k = \sum_{i} \left| \frac{\partial \mathcal{L}_{\text{pre-train}}}{\partial \theta_{k,i}} \odot \theta_{k,i} \right|,
\end{equation}
For simplicity we deonte the cost function as $\text{Cost}(\mathbf{m})$.
By constraining the total cost $\sum_{k \in \mathcal{S}} T_k$, we prioritize adapting layers with lower $T_k$ (i.e., those less critical to the pre-trained knowledge).
This ensures that adaptation focuses on "safe" regions of the network, reducing the risk of overwriting crucial pre-trained features.
Therefore, the cost function can be formulated as:
\begin{equation}
    \text{Cost}(\mathbf{m})=\sum_{k\in\mathcal{S}}\sum_{i}T_{k,i}\cdot \mathbf{m}_{k,i}.
\end{equation}

\noindent{\textbf{Taylor Approximation of the Task-Specific Loss Function.}}
We start by analyzing the sensitivity of the loss function $\mathcal{L}$ to the mask variable $\mathbf{m}$.
Assuming we have local smoothness around $\mathbf{m}=\mathbbm{1}$, then the loss can be approximated using a second-order Taylor expansion:
\begin{equation}\label{eq:loss}
    \mathcal{L}(\mathbf{m};\tilde{\bm{\theta}})\approx\mathcal{L}(\mathbbm{1};\tilde{\bm{\theta}})+\frac{1}{2}(\mathbbm{1}-\mathbf{m})^\top\mathbf{H}(\mathbbm{1}-\mathbf{m}),
\end{equation}
where $\tilde{\bm{\theta}}=\bm{\theta}+\bm{\phi}$ denotes the merged weights.
Here, the first-order term $\nabla\mathcal{L}(\mathbbm{1};\tilde{\bm{\theta}})^\top(\mathbf{m}-\mathbbm{1})=0$ due to the assumption that the model has converged to a local minima, where the gradient term is close to 0~\citep{frantar2023sparsegpt,wang2019eigendamage,NIPS1989_6c9882bb}.
As $\mathcal{L}(\mathbbm{1};\tilde{\bm{\theta}})$ is a constant, we can rewrite the optimization objective in Eq.~\ref{eq:loss} as follows:
\begin{equation}\label{eq:obj1}
    \underset{\mathbf{m}}{\argmin}\ \mathcal{L}(\mathbf{m})\approx\underset{\mathbf{m}}{\argmin}(\mathbbm{1}-\mathbf{m})^\top\mathbf{H}(\mathbbm{1}-\mathbf{m}).
\end{equation}
Eq.~\ref{eq:obj1} shows that the optimal mask is determined by the Hessian of the loss with respect to the mask variables, i.e. $\mathbf{H}=\mathbb{E}_{x\sim\mathcal{D}}[\nabla_{\mathbf{m}}^2\mathcal{L}(\mathbbm{1};\tilde{\bm{\theta}})]$.
Since computing the exact Hessian matrix is infeasible, we approximate the Hessian $\mathbf{H}$ with the empirical Fisher Information Matrix (FIM), which is defined as:
\begin{equation}
    \mathcal{I}(\mathbf{m}) = \mathbb{E}_{x \sim\mathcal{D}} \left[ \nabla_{\mathbf{m}}\mathcal{L}(\mathbbm{1})\nabla_{\mathbf{m}}\mathcal{L}(\mathbbm{1})^\top \right].
\end{equation}
\noindent{\textbf{Diagonal Approximation of the FIM.}} Assuming each layer $k$ contains $|\theta_k|$ parameters (including the weight parameters of both MHA and FFN components), then $\mathbf{m}_k$ can be seen as a vector of length $|\theta_k|$.
As $\mathbf{m}$ is applied across all $L$ layers, the full FIM $\mathcal{I}$ has $L^2|\theta_k|^2$ elements, making its computation and storage intractable for large values of $|\theta_k|$.
To address this challenge, we adopt a diagonal approximation of $\mathcal{I}$, reducing its complexity from $O(L^2|\theta_k|^2)$ to $O(L|\theta_k|)$.
This approximation is based on an assumption that cross-layer interactions can be neglected, since the off-diagonal terms $\mathcal{I}_{k,l}$ ($k \neq l$) are ignored.
Under this assumption, only the diagonal elements $\mathcal{I}_{k,k}$ are computed for each layer $k$, where:
\begin{equation}
    \mathcal{I}_{k,k} = \mathbb{E}_{x\sim\mathcal{D}}\left[ \frac{\partial \mathcal{L}}{\partial \mathbf{m}_k} \right]^2.
\end{equation}
This further simplifies Eq.~\ref{eq:obj1} as follows:
\begin{equation}\label{eq:obj2}
\begin{aligned}
    \underset{\mathbf{m}}{\argmin}\ \mathcal{L}(\mathbf{m}) 
    &\approx \underset{\mathbf{m}}{\argmin}\sum_{k=1}^L(\mathbbm{1}-\mathbf{m}_k)^2\mathcal{I}_{k,k} \\
    &= \underset{\mathbf{m}}{\argmin}\sum_{k=1}^L\sum_{i=1}^{|\theta_k|}(1-\mathbf{m}_{k,i})^2\mathcal{I}_{k,k,i}.
\end{aligned}
\end{equation}
Let $Z_k(\mathbf{m})=\{i:\mathbf{m}_{k,i}=0\}$. Since we restrict the possible mask values to either 0 or 1, the following can be derived from Eq.~\ref{eq:obj2}:
\begin{equation}
    \underset{\mathbf{m}}{\argmin}\ \mathcal{L}(\mathbf{m})\approx\underset{\mathbf{m}}{\argmin}\sum_{k}\sum_{i\in Z_k(\mathbf{m})}\mathcal{I}_{k,k,i}.
\end{equation}
Then the optimization objective in Eq.~\ref{eq:obj} is equivalent to minimizing the sum of layer-wise Fisher information of the unmasked parameters:
\begin{equation}\label{eq:obj3}
    \underset{\mathbf{m}}{\argmin}\sum_{k}\sum_{i\in Z_k(\mathbf{m})}\mathcal{I}_{k,k,i}\quad\text{s.t.}\quad\sum_{k}\sum_{i\in Z_k(\mathbf{m})}T_{k,i}\leq C.
\end{equation}

\subsection{Solving the Contrained Optimization Problem}
\noindent{\textbf{Determination Stage.}}
Within each transformer layer, the architecture consists of two primary components: a multi-head attention (MHA) module and a feed-forward network (FFN). We denote the mask variables for these components in layer $k$ as $\mathbf{m}^{\text{MHA}}_k$ and $\mathbf{m}^{\text{FFN}}_k$, respectively.
Suppose there are $N^{\text{MHA}}$ head mask variables and $N^{\text{FFN}}$ neuron mask variables.

The optimization problem in Eq.~\ref{eq:obj3} can be interpreted as follows: For each layer $k$, we seek a subset of unmasked parameters $\mathcal{M}_k$ that minimizes their total Fisher information on the adapted model $\mathcal{F}(\tilde{\bm{\theta})})$, while constraining their total Taylor importance (computed on the base model $\mathcal{F}(\bm{\theta})$) under a global budget $C$.

As the Fisher information and Taylor importance vary across individual parameters, Eq.~\ref{eq:obj3} becomes a dynamic programming problem which is memory inefficient. To reduce the 2-dimensional search space into a linear form, we employ a component-wise approximation within the same layer. Specifically, we compute the scores for MHA and FFN respectively, and then average these values to yield a single importance estimate per parameter.
This allows us to use a greedy solution (described in Algorithm ~\ref{alg:mask_search}).
The algorithm iteratively excludes the parameters with smallest Taylor importance until the budget $C$ is reached, while maximizing the cumulative Fisher information of the included parameters.

\begin{algorithm}[t]
\caption{Greedy Mask Search}\label{alg:mask_search}
\begin{algorithmic}[1]
\State \textbf{Input:} Budget $C$, Fisher score per-head $\{\tau_i^h\}_{i=1}^{N^{\text{MHA}}}$, Fisher score per-neuron (for FFN) $\{\tau_j^f\}_{j=1}^{N^{\text{FFN}}}$, Taylor score per-head $t_h$, Taylor score per-neuron $t_f$.
\State Initialize optimal Fisher loss $\mathcal{L}^*\leftarrow\infty$, optimal sets of unmasked indices $(\mathcal{M}_h^*,\mathcal{M}_f^*)\leftarrow(\emptyset,\emptyset)$, mask variables $(\mathbf{m}^{\text{MHA}}, \mathbf{m}^{\text{FFN}})\leftarrow(\mathbbm{1},\mathbbm{1})$
\For{$n=0$ to $N^{\text{MHA}}$}
    \State Compute cost for MHA module: $\hat{C}_h=n\cdot t_h$
    \If{$\hat{C}_h>C$}
        \State \textbf{continue} \Comment{Exceeds budget}
    \EndIf
    \State Remaining budget: $C_r=C-\hat{C}_h$
    \State Retained neurons: $f=\min\left( \max\left(0,\left\lfloor\dfrac{C_r}{t_f}\right\rfloor\right), N^{\text{FFN}} \right)$
    \State Select $n$ heads with smallest $\tau_i^h$: indices $\mathcal{P}_h$
    \State Select $f$ neurons with smallest $\tau_j^f$: indices $\mathcal{P}_f$
    \State Compute total loss: $\mathcal{L}=\sum_{i\in\mathcal{P}_h}\tau_i^h+\sum_{j\in\mathcal{P}_f}\tau_j^f$
    \If{$\mathcal{L}<\mathcal{L}^*$}
        \State $\mathcal{L}^*\leftarrow\mathcal{L}$, $(\mathcal{M}_h^*,\mathcal{M}_f^*)\leftarrow(\mathcal{P}_h,\mathcal{P}_f)$
    \EndIf
\EndFor
\State Apply masks: Set $\mathbf{m}^{\text{MHA}}[\mathcal{M}_h^*]=0$, $\mathbf{m}^{\text{FFN}}[\mathcal{M}_f^*]=0$
\State \textbf{Output:} Optimal mask $\mathbf{m}^*=(\mathbf{m}^{\text{MHA}},\mathbf{m}^{\text{FFN}})$
\end{algorithmic}
\end{algorithm}

\noindent{\textbf{Refinement Stage.}}
The component-wise approximation in determination stage decouples the selection of MHA heads and FFN neurons within each layer, thereby ignoring potential interactions between them. While efficient, this approximation may lead to suboptimal trade-offs between Taylor importance and Fisher information. To mitigate this, we propose a post-hoc refinement stage that jointly optimizes the masks for both components under the same global budget constraint. This refinement operates on the initial greedy solution as a warm start, enabling recovery of near-optimal masks with minimal computational overhead.

The refinement process is designed to iteratively adjust the selected MHA heads and FFN neurons while respecting the budget constraint.
Let $\mathcal{M}_h^*$ and $\mathcal{M}_f^*$ denote the sets of unmasked MHA heads and FFN neurons from the greedy solution. We define the refinement loss for a candidate mask pair $(\mathcal{M}_h, \mathcal{M}_f)$ as:
\begin{equation}
    \mathcal{L}(\mathcal{M}_h, \mathcal{M}_f) = \sum_{i \notin \mathcal{M}_h} \tau_i^h + \sum_{j \notin \mathcal{M}_f} \tau_j^f,\quad\text{s.t.}\quad\sum_{i \notin \mathcal{M}_h} t_h + \sum_{j \notin \mathcal{M}_f} t_f \leq C.
\end{equation}
The goal is to perturb $(\mathcal{M}_h^*, \mathcal{M}_f^*)$ to minimize $\mathcal{L}$ under the constraint.
Let $\mathcal{C}$ denote a joint candidate set containing all parameters (both masked and unmasked) at the same layer:
\begin{equation}
    \mathcal{C} = \left\{ (i, j) \mid i \in \mathcal{M}_h^*, j \in \mathcal{Q}_\text{FFN}\setminus\mathcal{M}_f^* \right\} \cup \left\{ (i, j) \mid i \in \mathcal{M}_f^*,j\in\mathcal{Q}_\text{MHA}\setminus\mathcal{M}_h^* \right\}.
\end{equation}
where $\mathcal{Q}_\text{MHA}$ and $\mathcal{Q}_\text{FFN}$ represent the complete sets of parameters in MHA and FFN modules, respectively.
For each candidate parameter $p \in \mathcal{C}$, compute the \textit{swap gain} if $p$ is masked and another parameter $q$ (of any component) is unmasked to compensate for the budget:
\begin{equation}
    \Delta \mathcal{L}_{p \rightarrow q} = \tau_p - \tau_q, \quad \Delta C_{p \rightarrow q} = t_p - t_q.
\end{equation}
A valid swap satisfies $\Delta C_{p \rightarrow q} \geq 0$ to preserve the budget constraint.
Then we select the swap with the largest $\Delta \mathcal{L}_{p \rightarrow q}$ (i.e., maximal reduction in total Fisher loss) to update $\mathcal{M}_h^*$, $\mathcal{M}_f^*$ and the remaining budget. Repeat this process until no improving swaps exist. The algorithm is implemented in Algorithm~\ref{alg:mask_refinement}.

The refinement stage approximates a single iteration of the Lagrange multiplier method, where swaps implicitly adjust the balance between Taylor importance (constraint) and Fisher information (objective). By restricting swaps to the vicinity of the initial greedy solution, it avoids the $O(|\tilde{\bm{\theta}}||\bm{\theta}|)$ complexity of full dynamic programming while recovering Pareto-improved solutions. 

\noindent{\textbf{Tuning Stage.}}
Since our goal is to use mask values to measure parameter importance, the initial binary masks are insufficient, as they fail to capture parameters that contribute marginally on their own but are collectively important. To address this limitation, we introduce a differentiable tuning stage that relaxes the binary masks into continuous values.
The layer-wise reconstruction objective is formulated as the residual activation difference:
\begin{equation}\label{eq:reconstruction}
    \argmin_{\mathbf{m}_k}\left\| \mathcal{O}(x; \mathbbm{1}) - \mathcal{O}(x'; \mathbf{m}_k) \right\|^2_2
\end{equation}
where $x'$ and $x$ are inputs to the layer with or without mask, $\mathcal{O}(x;\mathbf{m}_k)=x+l_k(x;\mathbf{m}_k)$ denotes the residual output of a mask-scaled layer, and $l_k$ indicates a MHA or FFN layer.
A detailed derivation is provided in Appendix \ref{app:tuning_stage_impl}.

\subsection{Dynamic Rank Selection}
We use Bayesian Optimization (BO) to efficiently determine the optimal rank for a given dataset.
BO constructs a probabilistic surrogate model to approximate the relationship between target hyperparameter $r$ and validation loss.
Based on this surrogate, an acquisition function iteratively proposes candidate $r$ values based on the past accuracy results, and focusing evaluation on values that are likely to improve performance.
Compared to grid and random search, BO reduces costly fine-tuning trials and discards suboptimal ranks early, making it efficient for expensive fine-tuning runs.

\begin{figure*}[h]
    \centering
    \includegraphics[width=\linewidth]{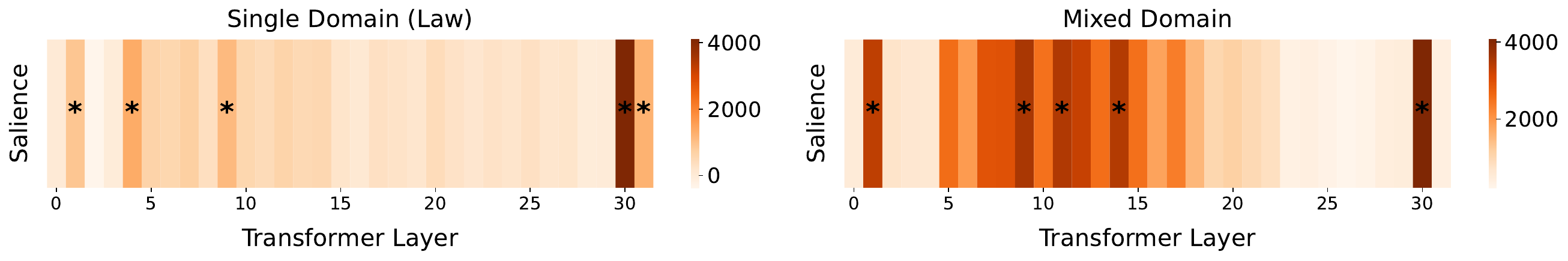}
    \caption{\small Saliency maps illustrating the importance of layers on the single-domain (Lawyer-Instruct) and mixed-domain/task (FLANv2) datasets. The base model is LLaMA2-7B. Black dots indicates the top-5 critical layers for the given task.}
    \label{fig:llama_imp}
\end{figure*}

\begin{table*}[h]
\small
\centering
\caption{\small Performance comparison of PEFT methods on LLaMA2-7B. Performance metrics include accuracy on MMLU (5-shot) and GSM8K, Pass@1/Pass@10 on HumanEval, and trainable parameter count (\% Param).
LoRA and its variants (AdaLoRA, DoRA and HydraLoRA) adopt the same parameter setting with $r=8$ and use a single A matrix, but differ in the number of B matrices.
Our method achieves the best results (at 8/8/20/16/24 layers respectively) across all 5 tasks with an average of 0.062\% training parameters compared to the full model.}
\setlength{\tabcolsep}{6pt}
\resizebox{0.9\textwidth}{!}{
\renewcommand\arraystretch{1.28}
\begin{tabular}{c|cccccc|c|c}
\bottomrule[1.5pt]
 & & & & \multicolumn{2}{c}{\textbf{HumanEval}} & & & \\
 \multirow{-2}{*}{\textbf{Schemes}} & \multirow{-2}{*}{\textbf{MMLU}}  &\multirow{-2}{*}{\textbf{Medical}}  & \multirow{-2}{*}{\textbf{Law}} & P@1&P@10&\multirow{-2}{*}{\textbf{GSM8K}}  & \multirow{-2}{*}{\textbf{\% Param}} & \multirow{-2}{*}{\textbf{B Num}} \\
 \toprule[0.75pt]
LLaMA2-7B~\citep{touvron2023llama} & 38.79 & 36.05 & 33.64 & 13.17 & 20.41 & 10.44 & - & -\\
Full Fine-Tuning & 49.91 & 46.76 & 46.22 & 20.24 & 32.93 & 25.69 & 100 & -\\
\midrule[0.75pt]
Prompt Tuning~\citep{lester2021power} & 39.97 & 37.46 & 34.88 & 13.59 & 21.62 & 13.25 & 0.001 & - \\
P-Tuning$_{(256)}$~\citep{liu2024gpt} & 41.02 & 39.85 & 36.64 & 13.53 & 21.20 & 15.50 & 0.193 & -\\
Prefix Tuning~\citep{li2021prefix} & 41.86 & 40.28 & 36.30 & 13.15 & 22.48 & 16.83 & 0.077 & -\\
LoRA$_{(r=8)}$~\citep{hu2021lora} & 45.88 & 46.76 & 37.16 & 14.57 & 29.88 & 13.72 & 0.062 & 1\\
AdaLoRA$_{(r=8)}$~\citep{zhang2023adalora} & 44.26 & 42.39 & 39.36 & 14.74 & 23.85 & 19.44 & 0.093 & 1\\
DoRA$_{(r=8)}$~\citep{liu2024dora} & 44.57 & 44.23 & 38.74 & 14.65 & 24.20 & 19.50 & 0.062 & 1 \\
HydraLoRA$_{(r=8)}$~\citep{tian2024hydralora} & 45.83 & 46.90 & 37.76 & 14.39 & 28.66 & 17.66 & 0.124 & 4 \\
\midrule[0.75pt]
\rowcolor{gray!20}\textbf{Ours}$_{(r=8)}$ & \textbf{46.48} & \textbf{49.15} & \textbf{39.14} & \textbf{14.82} & \textbf{31.71} & \textbf{20.09} & \textbf{0.062} & 4 \\
\bottomrule[1.5pt]
\end{tabular}}
\label{table:peft-result}
\end{table*}
\section{Experiments}\label{sec:exp}
\subsection{Experimental Setup}
\noindent{\textbf{Datasets and Benchmarks.}}
To investigate the effectiveness of our layer selection policy, we conduct experiments on both single- and multi-domain datasets. \textbf{Single domain} includes: \textit{General}, \textit{Medical}, \textit{Legal}, \textit{Code Generation}, \textit{Mathematics}. \textbf{Multi domain} includes \textit{FLANv2} and we evaluate the performance on BBH benchmark~\citep{suzgun2022challenging}.
Detailed descriptions of the datasets and benchmarks are provided in Appendix~\ref{app:datasets}.

\noindent{\textbf{Baselines.}}
To evaluate the adaptation performance on \model~-selected layers, we compare it with different PEFT methods: \textit{Full Fine-Tuning}, \textit{Prompt Tuning}~\citep{lester2021power}, \textit{P-Tuning}~\citep{liu2024gpt}, \textit{LoRA}~\citep{hu2021lora}, \textit{AdaLoRA}~\citep{zhang2023adalora}, \textit{DoRA}~\citep{liu2024dora}, \textit{HydraLoRA}~\citep{tian2024hydralora}.
To evaluate the layer selection policy, we compare \model~ with \textit{Random Selection} and \textit{Weight Norm Selection}~\citep{pan2024lisa}.
We further compare \model~ with two LoRA derivatives, \textit{LoraHub} \cite{huang2023lorahub} and \textit{LoRAMoE} \cite{dou2023loramoe}, which also utilize a routing mechanism to coordinate multiple LoRA experts.
Detailed descriptions of these baseline methods are provided in Appendix~\ref{app:baselines}.

\subsection{Main Results}
\noindent{\textbf{Implementation Details.}} 
For single-domain adaptation, we first fine-tune a fully adapted model on Databricks-Dolly-15K, where each adapter consists of one A matrix and four B matrices with rank set to 8. We apply \model~ on this model to select critical layers.
Then we fine-tune the sparsely adapted model on the target dataset.
For mixed-domain adaptation, we sample a small subset of FLANv2 (with a ratio of 1.25\%) for layer selection, and then fine-tune the model on a larger training set sampled from FLANv2 (with a ratio of 3\%). All the training tasks on running on A800-80G GPUs. Inference and evaluation tasks are running on A800-80G and A40-40G GPUs.

\noindent{\textbf{{Results on Single/Mixed Domain dataset.}}
The experimental results are presented in Table~\ref{table:peft-result} for fine-tuning performance evaluation and Table~\ref{table:lisresult} for layer selection policy evaluation.
These results demonstrate that \model~ consistently outperforms all competing approaches while reducing a large portion of trainable parameters.
Results in Table~\ref{table:layer} show that the optimal number of layers varies depending on the specific domain, with fewer layers generally performing better on MMLU and Medical benchmarks, while a moderate number of layers might be more effective for Law and GSM8K benchmarks.
Figure~\ref{fig:bbh} shows the results of mixed tasks and domains. The A/B configuration is 48/48 for LoraHub and LoRAMoE, 1/1 for LoRA, 1/6 for \model~ and HydraLoRA. Detailed results for each task in BBH evaluation are provided in Table~\ref{app:bbh_results}. The visualizations of the layer importance on single and mixed domain dataset is represented in Figure~\ref{fig:llama_imp}.

\noindent{\textbf{Dynamic Rank Selection.}}
Figure~\ref{fig:bo} presents the fine-tuning results of LLaMA2-7B under the different configurations of ranks and layers. The best performance is achieved at $r=4$.
We employed Bayesian optimization via Optuna~\citep{akiba2019optuna}, searching over integer values $r \in [2, 16]$ in 100 trials. We use Tree-structured Parzen Estimator (TPE) as the surrogate model, and choose the Expected Improvement (EI) acquisition function.

\begin{figure}[tb!]
    \centering
    \begin{minipage}[c]{0.48\textwidth}
        \centering
        \includegraphics[width=0.95\linewidth]{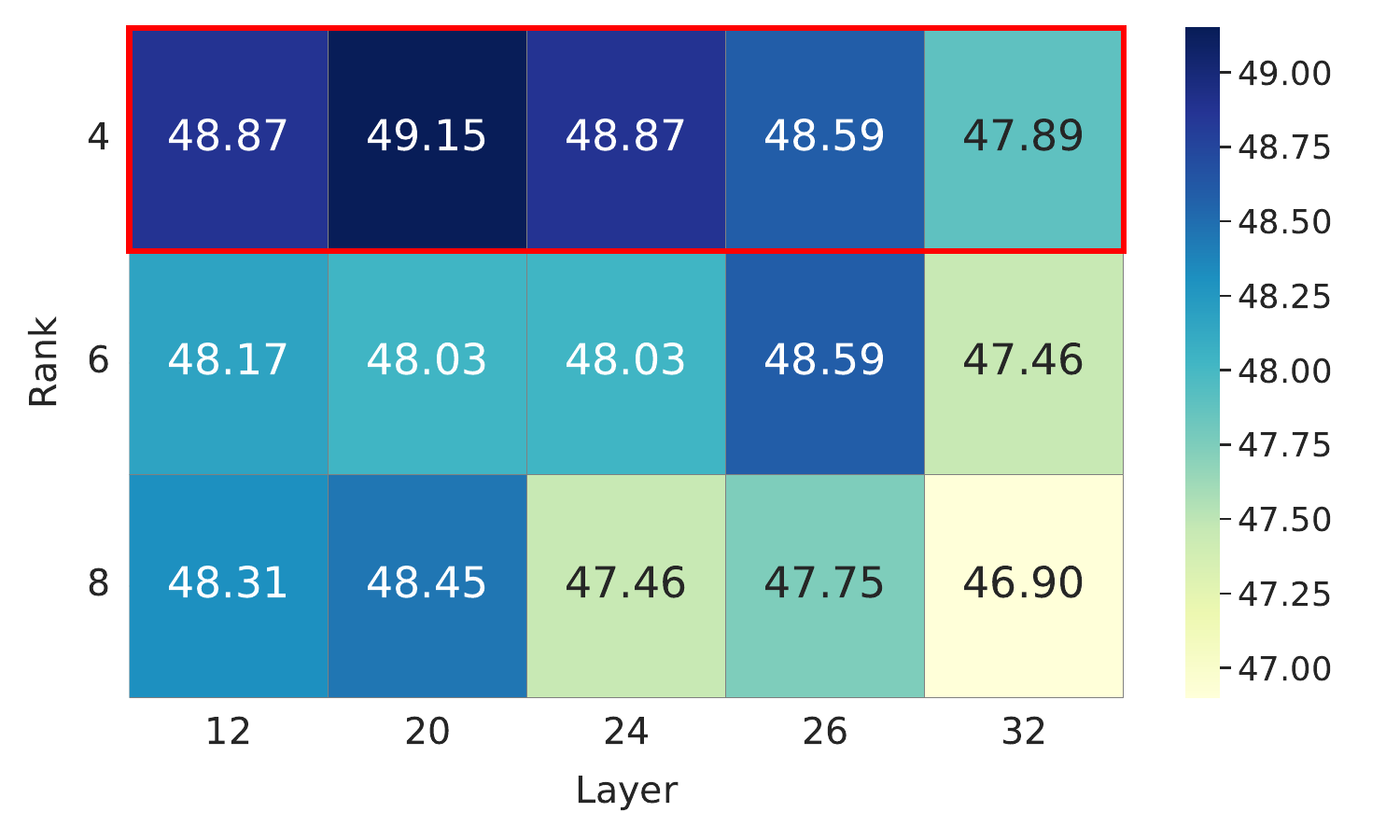}
        \caption{\small Dynamic rank selection results for the medical task on LLaMA2-7B. The optimal rank ($r=4$) is determined using Bayesian optimization on a full-layer adapted model (1A/4B).}
        \label{fig:bo}
    \end{minipage}
    \hspace{3mm}
    \begin{minipage}[c]{0.42\textwidth}
        \centering
        \includegraphics[width=0.95\linewidth]{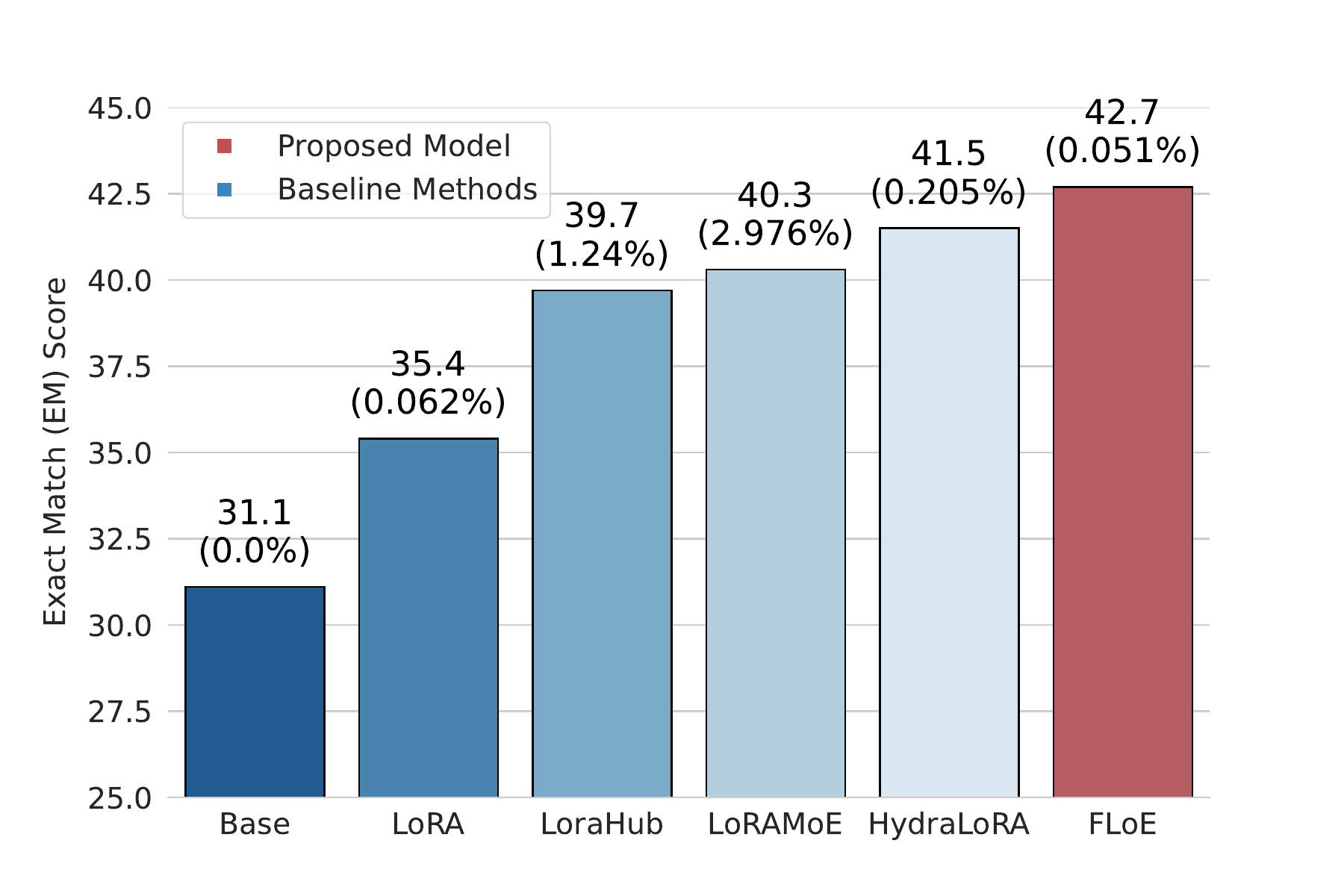}
        \caption{\small Mixed domain results evaluated on the BBH benchmark (3-shot) using LLaMA2-7B. \model~ achieves the highest Exact Match (EM) score with the lowest training parameter size.}
        \label{fig:bbh}
    \end{minipage}
\end{figure}
\begin{table*}[tb!]
\small
\centering
\caption{\small Performance comparison of different layer selection policies, including random selection (with a seed of 42), weight norm ($L_2$-norm of the pre-trained weight) and our proposed \model~ training across General and Medical domain on MMLU benchmark (5-shot) with LLaMA2-7B as the base model.}
\setlength{\tabcolsep}{8pt}
\resizebox{0.9\textwidth}{!}{
\begin{tabular}{c|cc>{\columncolor{gray!20}}c|cc>{\columncolor{gray!20}}c}
\toprule[1.5pt]
 & \multicolumn{3}{c}{\textbf{General}} & \multicolumn{3}{c}{\textbf{Medical}} \\
 \multirow{-2}{*}{\textbf{Layers}} & Random & WeightNorm & Ours (\model~) & Random & WeightNorm & Ours (\model~) \\
 \toprule[0.75pt]
32 & 45.83 & 45.83 & \textbf{45.83} & 46.90 & 46.90 & \textbf{46.90} \\
26 & 45.72 & 44.42 & \textbf{46.05} & 47.04 & 45.71 & \textbf{47.75} \\
24 & 46.15 & 44.86 & \textbf{46.24} & 47.46 & 46.13 & \textbf{47.46} \\
20 & 45.09 & 44.45 & \textbf{45.59} & 48.17 & 47.48 & \textbf{47.75} \\
16 & 45.58 & 45.02 & \textbf{45.61} & 47.46 & 46.89 & \textbf{47.89} \\
12 & 45.40 & 45.40 & \textbf{46.29} & 46.62 & 46.62 & \textbf{48.31} \\
8 & 46.29 & 45.56 & \textbf{46.48} & 46.48 & 45.75 & \textbf{49.15} \\
\bottomrule[1.5pt]
\end{tabular}}
\label{table:lisresult}
\end{table*}
\begin{table*}[tb!]
\small
\centering
\caption{\small Layer-specific fine-tuning using \model~ on LLaMA2-7B. We apply a 1A/4B adapter ($r=8$) for each selected layer. Best results per column are bolded.}
\setlength{\tabcolsep}{8pt}
\resizebox{0.8\textwidth}{!}{
\renewcommand\arraystretch{1.28}
\begin{tabular}{c|cccccc|c}
\bottomrule[1.5pt]
 & & & & \multicolumn{2}{c}{\textbf{HumanEval}} & & \\
 \multirow{-2}{*}{\textbf{Layers}} & \multirow{-2}{*}{\textbf{MMLU}}  &\multirow{-2}{*}{\textbf{Medical}}  & \multirow{-2}{*}{\textbf{Law}} & P@1&P@10&\multirow{-2}{*}{\textbf{GSM8K}}  & \multirow{-2}{*}{\textbf{\% Param}} \\
 \toprule[0.75pt]
32 & 45.83 & 46.90 & 37.76 & 14.39 & 28.66 & 17.66 & 0.124\\
26 & 46.05 & 47.75 & 38.07 & 14.45 & 27.44 & 19.94 & 0.101\\
24 & 46.24 & 47.46 & 37.76 & 14.76 & \textbf{31.71} & \textbf{20.09} & 0.093\\
20 & 45.59 & 47.75 & \textbf{39.14} & \textbf{14.82} & 31.10 & 18.62 & 0.078\\
16 & 45.61 & 48.03 & 38.00 & 13.90 & \textbf{31.71} & 18.65 & 0.062\\
12 & 46.29 & 48.45 & 37.89 & 13.78 & 29.27 &19.11& 0.047\\
8 & \textbf{46.48} & \textbf{49.15} & 36.80 & 13.48 & 29.88 & 17.91 & 0.031\\
\bottomrule[1.5pt]
\end{tabular}}
\label{table:layer}
\end{table*}

\noindent{\textbf{Results on Other Models Families.}}
To validate the generalization of \model~, we extended experiments to Gemma2-2B~\citep{team2024gemma} and Mistral-7B~\citep{jiang2024identifying}. As shown in Table~\ref{table:cross_model}, both models achieve optimal task performance when adapting 8 layers. Figure~\ref{fig:gemma_mistral_imp} visualizes layer importance distributions, showing that \model~ identifies middle layers for Mistral-7B and deep layers for Gemma2-2B as critical for adaptation.
\begin{table}[tb!]
\small
\centering
\caption{\small 
End-to-end inference latency by running the full MMLU benchmark (including 14,042 examples) on the fine-tuned Gemma2-2B and Mistral-7B models, the configuration for multi LoRA head is 1A/4B. Models are fine-tuned with Dolly-15K. Layer selection is performed using \model~.
}
\setlength{\tabcolsep}{8pt}
\resizebox{\textwidth}{!}{
\begin{tabular}{c|c|ccccccccc}
\toprule[1.5pt]
\multirow{2}{*}{\textbf{Models}} & \multirow{2}{*}{\textbf{Metrics}} & \multirow{2}{*}{\textbf{Single LoRA Head}} & \multicolumn{7}{c}{\textbf{Multi LoRA Heads}} \\
& & & \textbf{32}  & \textbf{28} & \textbf{24}  & \textbf{20} & \textbf{16} & \textbf{12} & \textbf{8} \\
\toprule[0.75pt]
 \multirow{2}{*}{Gemma2-2B} & \% Performance (Acc) & 51.25 & - & 51.56 & 51.73 & 51.67 & 51.12 & 51.30 & \textbf{51.99}$^{\uparrow 0.74}$ \\
 & Inference Lat. (s) & \textbf{1717.90} & - & 3832.37 & 3670.04 & 3368.81 & 3123.01 & 2777.69 & \textbf{2516.51}$^{\uparrow 798.61}$ \\
 \midrule[0.75pt]
 \multirow{2}{*}{Mistral-7B} & \% Performance (Acc) & 60.95 & 53.18 & 61.42 & 61.04 & 60.40 & 61.10 & 61.52 & \textbf{62.14}$^{\uparrow 1.19}$ \\
 & Inference Lat. (s) & \textbf{1462.27} & 4022.30 & 2055.37 & 1917.85 & 1758.16 & 1666.58 & 1636.02 & \textbf{1519.14}$^{\uparrow 56.87}$ \\
\bottomrule[1.5pt]
\end{tabular}}
\label{table:cross_model}
\end{table}

\noindent{\textbf{Inference Latency.}}
As we conduct experiments using a hydra architecture of LoRA, it inherently has more parameters than vanilla LoRA, we compare the latency of running inference on full-layer adapted models with single LoRA head (per adapter) and selected-layer adapted models with multiple LoRA heads (per adapter). Experiments are running on a single A40-40G GPU with a batch size of 16. While the full-layer LoRA-adapted models achieve minimal inference time, our \model~ layer selection strategy significantly optimizes latency for multi-head configurations.
As shown in Table~\ref{table:cross_model}, \model~ reduces inference latency by strategically limiting the number of adapted layers.

\begin{table}[tb!]
\small
\centering
\caption{\small Training runtime for full-layer and selected-layer adaptation. The \model~(Total) is the sum of Selected-Layer Adaptation and \model~ Layer Selection.}
\setlength{\tabcolsep}{8pt}
\resizebox{0.95\textwidth}{!}{
\begin{tabular}{c|cccc}
\toprule[1.5pt]
 & \textbf{Dolly-15K} & \textbf{Clinic-10K} & \textbf{Lawyer-Instruct} & \textbf{CodeAlpaca} \\
\midrule
Full-Layer Adaptation (Baseline) & 20105.08 & 20105.08 & 10137.51 & 45092.51 \\
Selected-Layer Adaptation & 16455.04 & 6697.16 & 6895.77 & 16197.23 \\
\model~ Layer Selection & 122.86 & 125.45 & 123.13 & 127.39 \\
\midrule
\model~ (Total) & \textbf{16577.90} (-3527.18) & \textbf{6822.61} (-7281.19) & \textbf{7018.90} (-3118.61) & \textbf{16324.62} (-28767.89) \\
\bottomrule[1.5pt]
\end{tabular}}
\label{table:layer_select_overhead}
\end{table}

\noindent{\textbf{Trade-off bewteen Training and Layer Selection.}}
Table~\ref{table:layer_select_overhead} shows the runtime of \model~ layer selection process.
The results show that the overall runtime including \model~ layer selection process and selected-layer adaptation is still lower than full-layer adaptation.

\section{Conclusion}
In this work, we first discuss the limitations of deploying MoE-based LoRA modules on all transformer layers indiscriminately, where domain interference significantly degrades performance across diverse tasks.
To address this, we propose \model~, a novel PEFT method that introduces two key innovations: Fisher information-aware layer selection and Bayesian optimization-driven dynamic rank allocation.
\model~ first selects critical layers via Fisher information, then determines the optimal rank in less than 100 trials via Bayesian optimization.
This dual mechanism not only mitigates domain interference by isolating task-specific features but also achieves superior performance with minimal parameter growth.
Our experiments demonstrate that \model~ offers a scalable and resource-efficient pathway for adapting LLMs to specialized domains, advancing the practical deployment of LLMs under constrained computational budgets. Broader impacts are discussed in Appendix~\ref{app:impact}.

\section*{Limitation}\label{sec:limitation}
While our proposed \model~ algorithm effectively identifies critical layers for MoE-based LoRA fine-tuing, its current implementation limits each LoRA module with a fixed number of low-rank experts.
Future work can explore dynamic expert allocation mechanisms, where both the selection of critical layers and the number of experts per layer are jointly optimized, enabling more granular control over model capacity allocation and better computational resource utilization.

{
\small
\bibliography{main}

\begin{thebibliography}{10}

\bibitem{akiba2019optuna}
Takuya Akiba, Shotaro Sano, Toshihiko Yanase, Takeru Ohta, and Masanori Koyama.
\newblock {O}ptuna: A next-generation hyperparameter optimization framework.
\newblock In {\em The 25th ACM SIGKDD International Conference on Knowledge Discovery \& Data Mining}, pages 2623--2631, 2019.

\bibitem{Lawyer-Instruct}
Alignment-Lab-AI.
\newblock Lawyer-instruct, 2024.

\bibitem{chalkidis2023lexfiles}
Ilias Chalkidis, Nicolas Garneau, Catalina Goanta, Daniel~Martin Katz, and Anders S{\o}gaard.
\newblock Lexfiles and legallama: Facilitating english multinational legal language model development.
\newblock {\em arXiv preprint arXiv:2305.07507}, 2023.

\bibitem{codealpaca}
Sahil Chaudhary.
\newblock Code alpaca: An instruction-following llama model for code generation.
\newblock \url{https://github.com/sahil280114/codealpaca}, 2023.

\bibitem{Humaneval}
Mark Chen, Jerry Tworek, Heewoo Jun, Qiming Yuan, Henrique Ponde de~Oliveira Pinto, Jared Kaplan, Harri Edwards, Yuri Burda, Nicholas Joseph, Greg Brockman, et~al.
\newblock Evaluating large language models trained on code.
\newblock {\em arXiv preprint arXiv:2107.03374}, 2021.

\bibitem{chen2023longlora}
Yukang Chen, Shengju Qian, Haotian Tang, Xin Lai, Zhijian Liu, Song Han, and Jiaya Jia.
\newblock Longlora: Efficient fine-tuning of long-context large language models.
\newblock {\em arXiv preprint arXiv:2309.12307}, 2023.

\bibitem{gsm8k}
Karl Cobbe, Vineet Kosaraju, Mohammad Bavarian, Mark Chen, Heewoo Jun, Lukasz Kaiser, Matthias Plappert, Jerry Tworek, Jacob Hilton, Reiichiro Nakano, Christopher Hesse, and John Schulman.
\newblock Training verifiers to solve math word problems.
\newblock {\em arXiv preprint arXiv:2110.14168}, 2021.

\bibitem{ding2023sparse}
Ning Ding, Xingtai Lv, Qiaosen Wang, Yulin Chen, Bowen Zhou, Zhiyuan Liu, and Maosong Sun.
\newblock Sparse low-rank adaptation of pre-trained language models.
\newblock {\em arXiv preprint arXiv:2311.11696}, 2023.

\bibitem{dou2023loramoe}
Shihan Dou, Enyu Zhou, Yan Liu, Songyang Gao, Jun Zhao, Wei Shen, Yuhao Zhou, Zhiheng Xi, Xiao Wang, Xiaoran Fan, et~al.
\newblock Loramoe: Alleviate world knowledge forgetting in large language models via moe-style plugin.
\newblock {\em arXiv preprint arXiv:2312.09979}, 2023.

\bibitem{elhoushi2024layer}
Mostafa Elhoushi, Akshat Shrivastava, Diana Liskovich, Basil Hosmer, Bram Wasti, Liangzhen Lai, Anas Mahmoud, Bilge Acun, Saurabh Agarwal, Ahmed Roman, et~al.
\newblock Layer skip: Enabling early exit inference and self-speculative decoding.
\newblock {\em arXiv preprint arXiv:2404.16710}, 2024.

\bibitem{frantar2023sparsegpt}
Elias Frantar and Dan Alistarh.
\newblock Sparsegpt: Massive language models can be accurately pruned in one-shot.
\newblock In {\em International Conference on Machine Learning}, pages 10323--10337. PMLR, 2023.

\bibitem{gao-etal-2025-mola}
Chongyang Gao, Kezhen Chen, Jinmeng Rao, Ruibo Liu, Baochen Sun, Yawen Zhang, Daiyi Peng, Xiaoyuan Guo, and Vs~Subrahmanian.
\newblock {M}o{LA}: {M}o{E} {L}o{RA} with layer-wise expert allocation.
\newblock In Luis Chiruzzo, Alan Ritter, and Lu~Wang, editors, {\em Findings of the Association for Computational Linguistics: NAACL 2025}, pages 5097--5112, Albuquerque, New Mexico, April 2025. Association for Computational Linguistics.

\bibitem{hayou2024lora}
Soufiane Hayou, Nikhil Ghosh, and Bin Yu.
\newblock Lora+: Efficient low rank adaptation of large models.
\newblock {\em arXiv preprint arXiv:2402.12354}, 2024.

\bibitem{hendrycks2020measuring}
Dan Hendrycks, Collin Burns, Steven Basart, Andy Zou, Mantas Mazeika, Dawn Song, and Jacob Steinhardt.
\newblock Measuring massive multitask language understanding.
\newblock {\em arXiv preprint arXiv:2009.03300}, 2020.

\bibitem{hu2021lora}
Edward~J Hu, Yelong Shen, Phillip Wallis, Zeyuan Allen-Zhu, Yuanzhi Li, Shean Wang, Lu~Wang, and Weizhu Chen.
\newblock Lora: Low-rank adaptation of large language models.
\newblock {\em arXiv preprint arXiv:2106.09685}, 2021.

\bibitem{huang2023lorahub}
Chengsong Huang, Qian Liu, Bill~Yuchen Lin, Tianyu Pang, Chao Du, and Min Lin.
\newblock Lorahub: Efficient cross-task generalization via dynamic lora composition.
\newblock {\em arXiv preprint arXiv:2307.13269}, 2023.

\bibitem{jacobs1991adaptive}
Robert~A Jacobs, Michael~I Jordan, Steven~J Nowlan, and Geoffrey~E Hinton.
\newblock Adaptive mixtures of local experts.
\newblock {\em Neural computation}, 3(1):79--87, 1991.

\bibitem{jiang2024identifying}
Fengqing Jiang.
\newblock Identifying and mitigating vulnerabilities in llm-integrated applications.
\newblock Master's thesis, University of Washington, 2024.

\bibitem{jiang2024mora}
Ting Jiang, Shaohan Huang, Shengyue Luo, Zihan Zhang, Haizhen Huang, Furu Wei, Weiwei Deng, Feng Sun, Qi~Zhang, Deqing Wang, et~al.
\newblock Mora: High-rank updating for parameter-efficient fine-tuning.
\newblock {\em arXiv preprint arXiv:2405.12130}, 2024.

\bibitem{kaplun2023less}
Gal Kaplun, Andrey Gurevich, Tal Swisa, Mazor David, Shai Shalev-Shwartz, and Eran Malach.
\newblock Less is more: Selective layer finetuning with subtuning.
\newblock {\em arXiv preprint arXiv:2302.06354}, 2023.

\bibitem{kopiczko2023vera}
Dawid~J Kopiczko, Tijmen Blankevoort, and Yuki~M Asano.
\newblock Vera: Vector-based random matrix adaptation.
\newblock {\em arXiv preprint arXiv:2310.11454}, 2023.

\bibitem{NIPS1989_6c9882bb}
Yann LeCun, John Denker, and Sara Solla.
\newblock Optimal brain damage.
\newblock In D.~Touretzky, editor, {\em Advances in Neural Information Processing Systems}, volume~2. Morgan-Kaufmann, 1989.

\bibitem{lee2022surgical}
Yoonho Lee, Annie~S Chen, Fahim Tajwar, Ananya Kumar, Huaxiu Yao, Percy Liang, and Chelsea Finn.
\newblock Surgical fine-tuning improves adaptation to distribution shifts.
\newblock {\em arXiv preprint arXiv:2210.11466}, 2022.

\bibitem{lei2023conditional}
Tao Lei, Junwen Bai, Siddhartha Brahma, Joshua Ainslie, Kenton Lee, Yanqi Zhou, Nan Du, Vincent Zhao, Yuexin Wu, Bo~Li, et~al.
\newblock Conditional adapters: Parameter-efficient transfer learning with fast inference.
\newblock {\em Advances in Neural Information Processing Systems}, 36:8152--8172, 2023.

\bibitem{lester2021power}
Brian Lester, Rami Al-Rfou, and Noah Constant.
\newblock The power of scale for parameter-efficient prompt tuning.
\newblock {\em arXiv preprint arXiv:2104.08691}, 2021.

\bibitem{li2021prefix}
Xiang~Lisa Li and Percy Liang.
\newblock Prefix-tuning: Optimizing continuous prompts for generation.
\newblock {\em arXiv preprint arXiv:2101.00190}, 2021.

\bibitem{li2023chatdoctor}
Yunxiang Li, Zihan Li, Kai Zhang, Ruilong Dan, Steve Jiang, and You Zhang.
\newblock Chatdoctor: A medical chat model fine-tuned on a large language model meta-ai (llama) using medical domain knowledge.
\newblock {\em Cureus}, 15(6), 2023.

\bibitem{liu2022few}
Haokun Liu, Derek Tam, Mohammed Muqeeth, Jay Mohta, Tenghao Huang, Mohit Bansal, and Colin~A Raffel.
\newblock Few-shot parameter-efficient fine-tuning is better and cheaper than in-context learning.
\newblock {\em Advances in Neural Information Processing Systems}, 35:1950--1965, 2022.

\bibitem{liu2024dora}
Shih-Yang Liu, Chien-Yi Wang, Hongxu Yin, Pavlo Molchanov, Yu-Chiang~Frank Wang, Kwang-Ting Cheng, and Min-Hung Chen.
\newblock Dora: Weight-decomposed low-rank adaptation.
\newblock {\em arXiv preprint arXiv:2402.09353}, 2024.

\bibitem{liu2021p}
Xiao Liu, Kaixuan Ji, Yicheng Fu, Weng~Lam Tam, Zhengxiao Du, Zhilin Yang, and Jie Tang.
\newblock P-tuning v2: Prompt tuning can be comparable to fine-tuning universally across scales and tasks.
\newblock {\em arXiv preprint arXiv:2110.07602}, 2021.

\bibitem{liu2024gpt}
Xiao Liu, Yanan Zheng, Zhengxiao Du, Ming Ding, Yujie Qian, Zhilin Yang, and Jie Tang.
\newblock Gpt understands, too.
\newblock {\em AI Open}, 5:208--215, 2024.

\bibitem{liu2024alora}
Zequan Liu, Jiawen Lyn, Wei Zhu, Xing Tian, and Yvette Graham.
\newblock Alora: Allocating low-rank adaptation for fine-tuning large language models.
\newblock {\em arXiv preprint arXiv:2403.16187}, 2024.

\bibitem{longpre2023flan}
Shayne Longpre, Le~Hou, Tu~Vu, Albert Webson, Hyung~Won Chung, Yi~Tay, Denny Zhou, Quoc~V Le, Barret Zoph, Jason Wei, et~al.
\newblock The flan collection: Designing data and methods for effective instruction tuning.
\newblock In {\em International Conference on Machine Learning}, pages 22631--22648. PMLR, 2023.

\bibitem{ma2023llm}
Xinyin Ma, Gongfan Fang, and Xinchao Wang.
\newblock Llm-pruner: On the structural pruning of large language models.
\newblock {\em Advances in neural information processing systems}, 36:21702--21720, 2023.

\bibitem{meng2404pissa}
Fanxu Meng, Zhaohui Wang, and Muhan Zhang.
\newblock Pissa: principal singular values and singular vectors adaptation of large language models.
\newblock {\em arXiv preprint arXiv:2404.02948}, 2024.

\bibitem{mike2023free}
C~Mike, H~Matt, M~Ankit, X~Jianwei, W~Jun, S~Sam, G~Ali, W~Patrick, Z~Matei, and X~Reynold.
\newblock Free dolly: Introducing the world’s first truly open instruction-tuned llm, 2023.

\bibitem{page2024multi}
Lucas Page-Caccia, Edoardo~Maria Ponti, Zhan Su, Matheus Pereira, Nicolas Le~Roux, and Alessandro Sordoni.
\newblock Multi-head adapter routing for cross-task generalization.
\newblock {\em Advances in Neural Information Processing Systems}, 36, 2024.

\bibitem{pan2024lisa}
Rui Pan, Xiang Liu, Shizhe Diao, Renjie Pi, Jipeng Zhang, Chi Han, and Tong Zhang.
\newblock Lisa: Layerwise importance sampling for memory-efficient large language model fine-tuning.
\newblock {\em arXiv preprint arXiv:2403.17919}, 2024.

\bibitem{renduchintala2023tied}
Adithya Renduchintala, Tugrul Konuk, and Oleksii Kuchaiev.
\newblock Tied-lora: Enhacing parameter efficiency of lora with weight tying.
\newblock {\em arXiv preprint arXiv:2311.09578}, 2023.

\bibitem{sajjad2023effect}
Hassan Sajjad, Fahim Dalvi, Nadir Durrani, and Preslav Nakov.
\newblock On the effect of dropping layers of pre-trained transformer models.
\newblock {\em Computer Speech \& Language}, 77:101429, 2023.

\bibitem{shazeer2017outrageously}
Noam Shazeer, Azalia Mirhoseini, Krzysztof Maziarz, Andy Davis, Quoc Le, Geoffrey Hinton, and Jeff Dean.
\newblock Outrageously large neural networks: The sparsely-gated mixture-of-experts layer.
\newblock {\em arXiv preprint arXiv:1701.06538}, 2017.

\bibitem{suzgun2022challenging}
Mirac Suzgun, Nathan Scales, Nathanael Sch{\"a}rli, Sebastian Gehrmann, Yi~Tay, Hyung~Won Chung, Aakanksha Chowdhery, Quoc~V Le, Ed~H Chi, Denny Zhou, , and Jason Wei.
\newblock Challenging big-bench tasks and whether chain-of-thought can solve them.
\newblock {\em arXiv preprint arXiv:2210.09261}, 2022.

\bibitem{team2024gemma}
Gemma Team, Morgane Riviere, Shreya Pathak, Pier~Giuseppe Sessa, Cassidy Hardin, Surya Bhupatiraju, L{\'e}onard Hussenot, Thomas Mesnard, Bobak Shahriari, Alexandre Ram{\'e}, et~al.
\newblock Gemma 2: Improving open language models at a practical size.
\newblock {\em arXiv preprint arXiv:2408.00118}, 2024.

\bibitem{tian2024hydralora}
Chunlin Tian, Zhan Shi, Zhijiang Guo, Li~Li, and Chengzhong Xu.
\newblock Hydralora: An asymmetric lora architecture for efficient fine-tuning.
\newblock {\em arXiv preprint arXiv:2404.19245}, 2024.

\bibitem{touvron2023llama}
Hugo Touvron, Louis Martin, Kevin Stone, Peter Albert, Amjad Almahairi, Yasmine Babaei, Nikolay Bashlykov, Soumya Batra, Prajjwal Bhargava, Shruti Bhosale, et~al.
\newblock Llama 2: Open foundation and fine-tuned chat models.
\newblock {\em arXiv preprint arXiv:2307.09288}, 2023.

\bibitem{valipour2022dylora}
Mojtaba Valipour, Mehdi Rezagholizadeh, Ivan Kobyzev, and Ali Ghodsi.
\newblock Dylora: Parameter efficient tuning of pre-trained models using dynamic search-free low-rank adaptation.
\newblock {\em arXiv preprint arXiv:2210.07558}, 2022.

\bibitem{wang2019eigendamage}
Chaoqi Wang, Roger Grosse, Sanja Fidler, and Guodong Zhang.
\newblock Eigendamage: Structured pruning in the kronecker-factored eigenbasis.
\newblock In {\em International conference on machine learning}, pages 6566--6575. PMLR, 2019.

\bibitem{wang2022adamix}
Yaqing Wang, Subhabrata Mukherjee, Xiaodong Liu, Jing Gao, Ahmed~Hassan Awadallah, and Jianfeng Gao.
\newblock Adamix: Mixture-of-adapter for parameter-efficient tuning of large language models.
\newblock {\em arXiv preprint arXiv:2205.12410}, 1(2):4, 2022.

\bibitem{flanv2}
Jason Wei, Maarten Bosma, Vincent~Y. Zhao, Kelvin Guu, Adams~Wei Yu, Brian Lester, Nan Du, Andrew~M. Dai, and Quoc~V. Le.
\newblock Finetuned language models are zero-shot learners.
\newblock In {\em The Tenth International Conference on Learning Representations, {ICLR} 2022, Virtual Event, April 25-29, 2022}. OpenReview.net, 2022.

\bibitem{zhang2023crash}
Kaiyan Zhang, Ning Ding, Biqing Qi, Xuekai Zhu, Xinwei Long, and Bowen Zhou.
\newblock Crash: Clustering, removing, and sharing enhance fine-tuning without full large language model.
\newblock {\em arXiv preprint arXiv:2310.15477}, 2023.

\bibitem{zhang2023lora}
Longteng Zhang, Lin Zhang, Shaohuai Shi, Xiaowen Chu, and Bo~Li.
\newblock Lora-fa: Memory-efficient low-rank adaptation for large language models fine-tuning.
\newblock {\em arXiv preprint arXiv:2308.03303}, 2023.

\bibitem{zhang2023adalora}
Qingru Zhang, Minshuo Chen, Alexander Bukharin, Nikos Karampatziakis, Pengcheng He, Yu~Cheng, Weizhu Chen, and Tuo Zhao.
\newblock Adalora: Adaptive budget allocation for parameter-efficient fine-tuning.
\newblock {\em arXiv preprint arXiv:2303.10512}, 2023.

\end{thebibliography}
\bibliographystyle{plain}
}

\appendix
\section{Datasets and Baselines}
\subsection{Datasets}\label{app:datasets}
\textbf{Single domain} includes:
\begin{itemize}[topsep=0pt,parsep=0pt,partopsep=0pt,leftmargin=2em]
    \item \textit{General}: we fine-tune with Databricks-Dolly-15K dataset~\citep{mike2023free} for general knowledge mastering and evaluate with all tasks in MMLU~\citep{hendrycks2020measuring}.
    \item \textit{Medical}: we fine-tune with GenMedGPT and Clinic-10K from ChatDoctor~\citep{li2023chatdoctor} for medical applications and evaluate with 3 medical tasks in MMLU, including \textit{clinical knowledge}, \textit{professional medicine} and \textit{college medicine}.
    \item \textit{Legal}: we fine-tune with Lawyer-Instruct \citep{Lawyer-Instruct} and US-Terms \citep{chalkidis2023lexfiles} for legal applications and evaluate with 3 legal tasks in MMLU, including \textit{jurisprudence}, \textit{international law} and \textit{professional law}.
    \item \textit{Code Generation}: we fine-tuned with CodeAlpaca~\citep{codealpaca} and evaluate with HumanEval~\citep{Humaneval}.
    \item \textit{Mathematics}: we fine-tune with the training split of GSM8K~\citep{gsm8k} for mathematical reasoning and evaluate with the test split.
\end{itemize}
\textbf{Multi domain} includes:
\begin{itemize}[topsep=0pt,parsep=0pt,partopsep=0pt,leftmargin=2em]
    \item \textit{FLANv2}: we construct the training dataset by sampling equal-proportion subsets from each of the 46 tasks in FLANv2~\cite{flanv2} and evaluate on BBH benchmark~\citep{suzgun2022challenging}.
\end{itemize}

\subsection{Baselines}\label{app:baselines}
Baselines for \textbf{PEFT}:
\begin{itemize}[topsep=0pt,parsep=0pt,partopsep=0pt,leftmargin=2em]
    \item \textit{Full Fine-Tuning}: the default adaptation strategy involves initializing the model with pre-trained weights and updating all parameters via gradient descent. The number of trainable parameters equals the number of pretrained parameters.
    \item \textit{Prompt Tuning}~\citep{lester2021power}: adds manually-designed task-specific prompts to the input. The fine-tuning process only updates the prompt parameters while keeping the pre-trained parameters frozen.
    \item \textit{P-Tuning}~\citep{liu2024gpt}: a prompt adds learnable prompt tokens to the input, optimized by a prompt encoder to find a better prompt. The prompt tokens can be added anywhere in the input sequence.
    \item \textit{LoRA}~\citep{hu2021lora}: decomposes weight updates into low-rank matrices, enabling efficient adaptation with significantly fewer trainable parameters while preserving model performance.
    \item \textit{AdaLoRA}~\citep{zhang2023adalora}: dynamically allocates trainable parameters across weight matrices and layers, prioritizing important components instead of uniformly distributing resources as in LoRA.
    \item \textit{DoRA}~\citep{liu2024dora}: decomposes pre-trained weights into magnitude and direction components, and apply LoRA on the direction component.
    \item \textit{HydraLoRA}~\citep{tian2024hydralora}: incorporating asymmetric LoRA adapters across all layers.
\end{itemize}
Baselines for \textbf{layer selection policy}:
\begin{itemize}[topsep=0pt,parsep=0pt,partopsep=0pt,leftmargin=2em]
    \item \textit{Random Selection}: layers are chosen uniformly at random during training (with a fixed random seed 42), serving as a naive baseline to assess the necessity of structured selection.
    \item \textit{Weight Norm Selection}~\citep{pan2024lisa}: layers are prioritized based on the $L_2$-norm of their weights, under the hypothesis that layers with larger norms contribute more significantly to the model's predictions.
\end{itemize}
Baselines for \textbf{multi-LoRA architecture}:
\begin{itemize}[topsep=0pt,parsep=0pt,partopsep=0pt,leftmargin=2em]
    \item \textit{LoRAMoE}~\cite{dou2023loramoe}: combines lightweight experts (LoRA) with MoE architecture for high efficiency, generalizing to new tasks without prior knowledge.
    \item \textit{LoraHub}~\cite{huang2023lorahub}: employs black-box optimization to learn weights of 20 randomly selected LoRAs for new tasks, using weighted averaging without needing gradient calculations.
\end{itemize}

\section{Hyperparameter Settings}
\begin{table}[h]
\small
\centering
\caption{\small Hyperparameter settings for our experiments and LoRA-based baseline methods. MSL indicates the max sequence length, BSZ indicates the batch size.
All methods use the AdamW optimizer. The weight decay is set to 0.
}
\setlength{\tabcolsep}{8pt}
\resizebox{\textwidth}{!}{
\begin{tabular}{c|cccccccccc}
\toprule[1.5pt]
\textbf{Parameter} & Rank & $\alpha$ & Dropout & MSL & Warmup Ratio & BSZ & Epochs & LR & Seed & Where \\
\midrule
\textbf{\model~} & 8 & 32 & 0.05 & 512 & 0.03 & 1 & 1 & 2e-4 & 42 & Up,Down,Gate \\
\textbf{HydraLoRA} & 8 & 32 & 0.05 & 1024 & 0.03 & 2 & 1 & 2e-4 & 41 & Up,Down,Gate \\
\textbf{LoRA} & 8 & 32 & 0.05 & 1024 & 0.03 & 2 & 1 & 2e-4 & 41 & Up,Down,Gate \\
\textbf{AdaLoRA} & 8 & 16 & 0.05 & 512 & 0.1 & 2 & 3 & 1e-3 & 41 & Up,Down,Gate\\
\textbf{DoRA} & 8 & 32 & 0.05 & 1024 & 0.03 & 2 & 1.5 & 2e-4 & 41 & Q,K,V,Up,Down \\
\bottomrule[1.5pt]
\end{tabular}}
\label{table:ft-args}
\end{table}

\section{Ablation Study}
The \model~ layer selection algorithm comprises three steps: (i) determination stage for coarse-grained binary mask value initialization, (ii) refinement stage implementing a swapping protocol that exchanges mask values between each pair of masked and unmasked parameters within the same layer, and (iii) tuning stage to relax the binary mask values into continuous values using a reconstruction objective.
To validate the necessity of refinement stage, we compare the model performance with and without the refinement stage.
We first fine-tune a full-layer LLaMA2-7B model with Databricks-Dolly-15K, and using \model~ to get the layer importance rankings which is shown in Figure~\ref{fig:ablation_rankings}.
Empirical results reveal that disabling this stage leads to a performance degradation as shown in Table~\ref{table:ablation_refinement_llama}.

\begin{figure*}[h]
    \centering
    \includegraphics[width=0.65\linewidth]{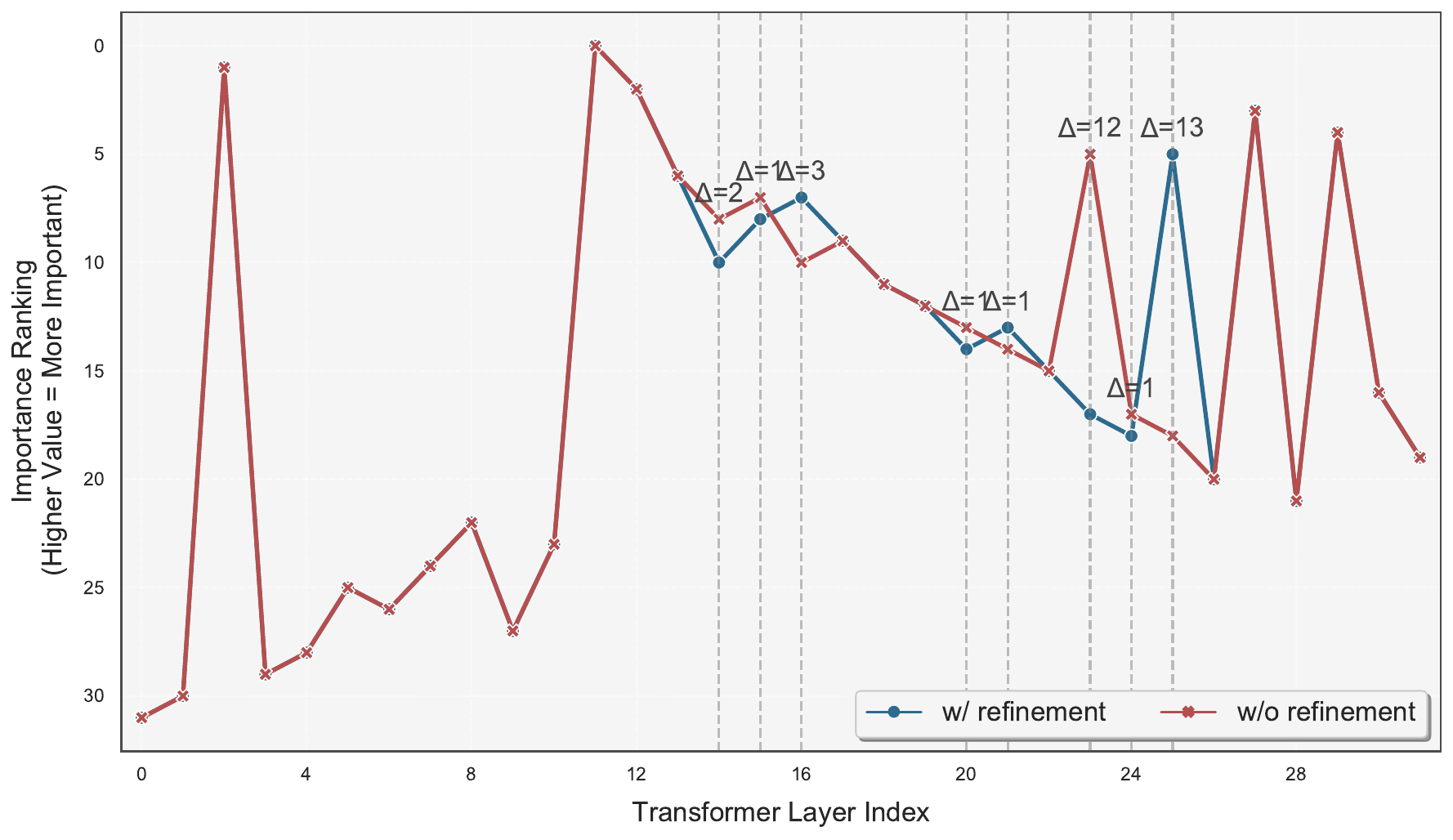}
    \caption{\small Layer importance rankings with or without refinement stage.
    The difference starts at the rank 14 (i.e. the former 15 layers. The 14-th layer is layer 10 in the w/o refinement rankings and layer 8 in the w/ refinement rankings) and ends at the rank 16 (i.e. the former 17 layers has the same set of layer indexes).
    Similarly, we only need to evaluate the former 21, 24 and 25 layers in the subsequent rankings.
    Both rankings and layer indexes start at 0. The range is 0-31.
    }
    \label{fig:ablation_rankings}
\end{figure*}

\begin{table}[h]
\small
\centering
\caption{\small Evaluation results on MMLU with the former 15, 16, 21, 24, 25 layers fine-tuned (in the ranking order).}
\setlength{\tabcolsep}{8pt}
\resizebox{0.75\textwidth}{!}{
\begin{tabular}{c|ccccc}
\toprule[1.5pt]
\textbf{Fine-tuned Layers} & \textbf{15} & \textbf{16} & \textbf{21} & \textbf{24} & \textbf{25} \\
\midrule
w/o refinement & 45.11 & 45.14 & 45.52 & \textbf{45.48} & 45.61 \\
w/ refinement & \textbf{45.36$^{\uparrow{0.25}}$} & \textbf{45.33$^{\uparrow{0.19}}$} & \textbf{45.56$^{\uparrow{0.04}}$} & 45.43$^{\downarrow{0.05}}$ & \textbf{45.66}$^{\uparrow{0.05}}$ \\
\bottomrule[1.5pt]
\end{tabular}}
\label{table:ablation_refinement_llama}
\end{table}

\section{Analysis for Layer Selection}
As \model~ layer selection algorithm need to run on an already fine-tuned model, we decrease the preparation overhead via reducing the size and category of training data.
For single domain fine-tuning, we first fine-tune a model on a general knowledge dataset, then run \model~ on that model. If this general dataset is too large, we reduce its size by randomly selecting a partial of the training data.
Note that we can use the selection results for other single domain fine-tuning including medical, law, code and mathematics.

\noindent{\textbf{Different Size of Sample Dataset.}}
To investigate the impact of training data quantity on layer selection, we conduct a comparative study using MoE-based LoRA for LLaMA2-7B fine-tuning. We evaluate layer importance rankings across three training scenarios: full-data (100\%), moderately reduced (70\%), and substantially reduced (30\%) subsets of the Databricks-Dolly-15K dataset.

Figure~\ref{fig:llama_imp_diff_size} illustrates the stability of layer rankings under these varying data regimes. Our analysis reveals that the refinement stage exhibits minimal sensitivity to training data quantity in terms of rank consistency. Across all three configurations, layers 0-2 and layers 10-14 demonstrate strong agreement in stability rankings. However, notable discrepancies emerge in the final layers (22-28), suggesting that deeper layers show higher variability in importance assessment when trained with reduced data.
This observation indicates that for a safe consideration, we should choose at least 50\% layers for subsequent fine-tuning on the target dataset. For a more careful layer selection, we should enlarge the sample dataset size.
\begin{figure*}[h]
    \centering
    \includegraphics[width=\linewidth]{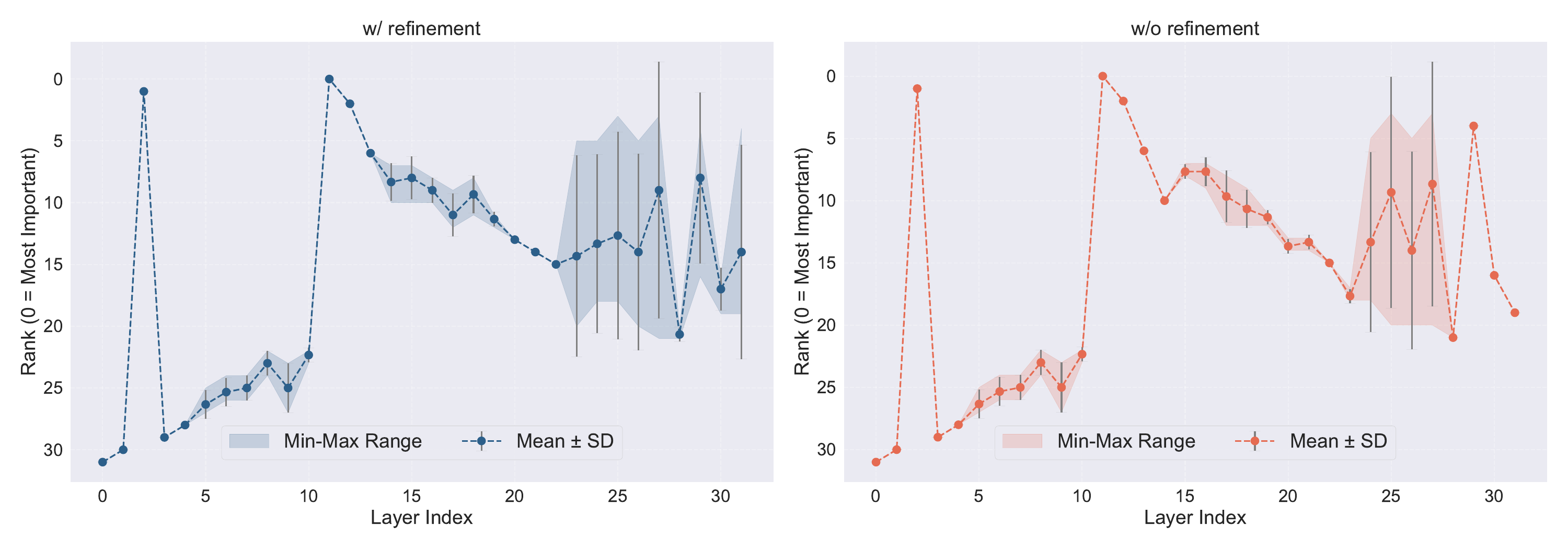}
    \caption{\small Rank stability comparison of layers with (left) and without (right) refinement stage. The plots show layer importance rankings (lower rank indicate higher importance) across three training data ratios (100\%, 70\%, 30\%). The shaded regions depict min-max ranges of the rank on different ratios.}
    \label{fig:llama_imp_diff_size}
\end{figure*}

\noindent{\textbf{Fine-tune on Different Target Datasets.}} After ranking layers on the sample dataset, we apply the ranking results to selectively fine-tune the base model on the target datasets. Note that the target dataset can be different from the sample dataset.
To evaluate the \textbf{generalization} capability of \model~, we compare its fine-tuning performance with two single domain datasets (Clinic-10K and CodeAlpaca). For each dataset, we implement two layer selection strategies: one using the general knowledge dataset (Databricks-Dolly-15K) as the sample dataset, and the other using the target dataset itself as the sample dataset.
Figure~\ref{fig:generalization} shows that although the misalignment between the sample and target datasets has little impact on the best achievable performance, it may affect the number of layers required to reach that performance.
\begin{figure*}[h]
    \centering
    \includegraphics[width=\linewidth]{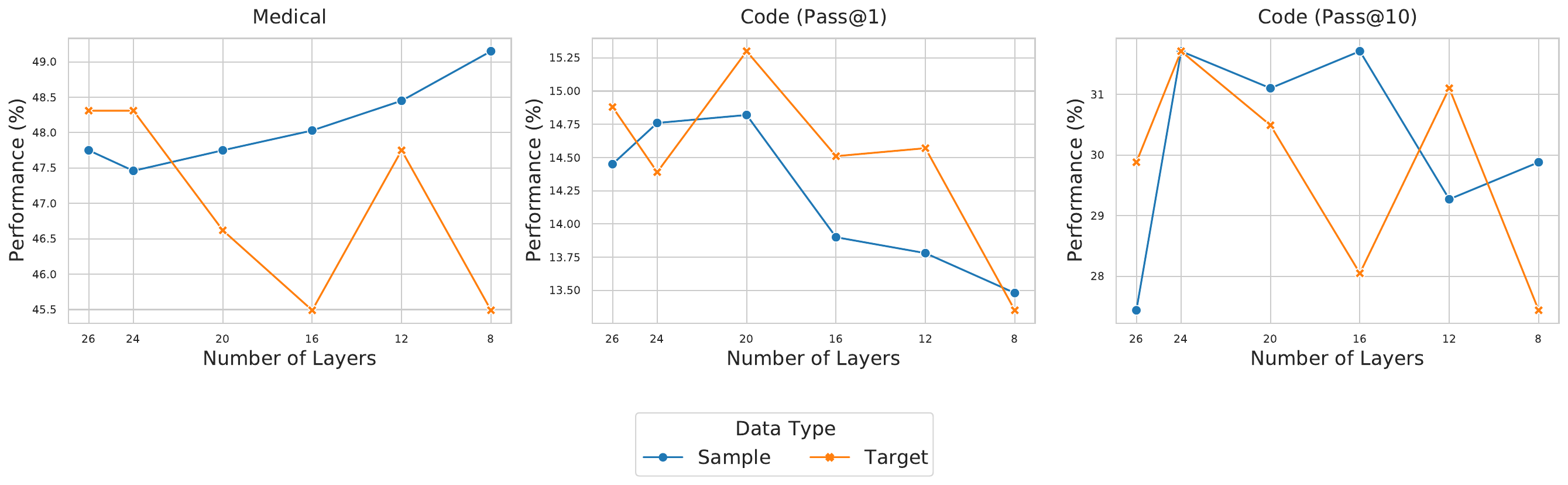}
    \caption{\small Illustration of performance under different layer importance rankings.}
    \label{fig:generalization}
\end{figure*}

\section{Proof of Eq.~\ref{eq:reconstruction}}\label{app:tuning_stage_impl}
The tuning stage is designed to minimize the layer-wise reconstruction error between the masked and unmasked versions of the adapted model.
The goal is to optimize $\mathbf{m}_k$ such that the squared $L_2$-norm of the residual activation difference is minimized:
\begin{equation}\label{app:rec-obj}
    \argmin_{\tilde{\mathbf{m}}_k} \left\| x' + l_k(x'; \mathbf{m}_k) - \left( x + l_k(x; \mathbbm{1}) \right) \right\|^2_2.
\end{equation}
Let $\mathbf{W}_i$ represent the $i$-th head or neuron weight of layer $l_k$. The output of $l_k(\cdot;\mathbf{m}_k)$ is:
\begin{equation}
    l_k(x; \mathbf{m}_k) = \sum_{i=1}^{N} \mathbf{m}_{k,i} \mathbf{W}_i x,
\end{equation}
where $N$ denote the number of heads (for MHA) or neurons (for FFN). Let $\Delta x=x'-x$, then the residual difference in Eq.~\ref{app:rec-obj} becomes:
\begin{align}
    x' + \sum_{i=1}^{N} \mathbf{m}_{k,i} \mathbf{W}_i x' - x - \sum_{i=1}^{N} \mathbbm{1}_i \mathbf{W}_i x
    &=\Delta x + \sum_{i=1}^{N} \mathbf{m}_{k,i} \mathbf{W}_i(x+\Delta x) - \sum_{i=1}^{N}\mathbbm{1}_i\mathbf{W}_i x\\ \nonumber
    &=\sum_{i=1}^N(\mathbf{m}_{k,i}-\mathbbm{1}_i)\mathbf{W}_ix + (\sum_{i=1}^{N} \mathbf{m}_{k,i}\mathbf{W}_i+1)\Delta x.
\end{align}
Let $\mathbf{A}=[\mathbf{W}_1x,\ldots,\mathbf{W}_Nx]$, $\mathbf{b} = -(\sum_{i=1}^{N} \mathbf{m}_{k,i}\mathbf{W}_i+1)\Delta x$, and $\mathbf{u}_k = \mathbf{m}_k - \mathbbm{1}$. Then Eq.~\ref{app:rec-obj} can be transformed into a linear least-square problem:
\begin{equation}\label{app:eq-lst}
    \argmin_{\mathbf{u}_k} \left\| \mathbf{A} \mathbf{u}_k - \mathbf{b} \right\|^2_2.
\end{equation}
Assume $\mathbf{A}^\top \mathbf{A}$ is invertible.
The closed-form solution to Eq.~\ref{app:eq-lst} is given by:
\begin{equation}
\mathbf{u}_k^* = (\mathbf{A}^\top \mathbf{A})^{-1} \mathbf{A}^\top \mathbf{b}.
\end{equation}
Finally, the optimal mask can be recovered by:
\begin{equation}
\mathbf{m}_k^* = \mathbf{u}_k^* + \mathbbm{1}.
\end{equation}

\section{More Results}\label{app:more_results}
\begin{algorithm}[h]
\caption{Joint Mask Refinement}\label{alg:mask_refinement} 
\begin{algorithmic}[1]
\State \textbf{Input:} Initial masks $\mathcal{M}_h^*, \mathcal{M}_f^*$, budget $C$, Fisher/Taylor scores.
\State Compute initial loss $\mathcal{L}^*$ and used budget $C_{\text{used}} = C - \left( \sum_{i \notin \mathcal{M}_h^*} t_h + \sum_{j \notin \mathcal{M}_f^*} t_f \right)$
\While{improvement}
    \State Generate all valid swaps $(p, q)$ where $\Delta C_{p \rightarrow q} \geq 0$
    \State Select swap $(p^*, q^*) = \arg\max_{(p,q)} \Delta \mathcal{L}_{p \rightarrow q}$
    \If{$\Delta \mathcal{L}_{p^* \rightarrow q^*} > 0$}  
        \State Update $\mathcal{M}_h^*$, $\mathcal{M}_f^*$, $\mathcal{L}^* \leftarrow \mathcal{L}^* - \Delta \mathcal{L}_{p^* \rightarrow q^*}$
        \State Update $C_{\text{used}} \leftarrow C_{\text{used}} + \Delta C_{p^* \rightarrow q^*}$
    \Else
        \State \textbf{break}
    \EndIf
\EndWhile
\State \textbf{Output:} Refined masks $\mathcal{M}_h^*, \mathcal{M}_f^*$
\end{algorithmic}
\end{algorithm}

\begin{table*}[h]
\centering
\caption{Detailed results of BBH evaluation with LLaMA2-7B as the base LLM (3-shot).}
\resizebox{0.95\linewidth}{!}{
\begin{tabular}{c|ccc>{\columncolor{gray!20}}c>{\columncolor{gray!20}}c>{\columncolor{gray!20}}c>{\columncolor{gray!20}}c>{\columncolor{gray!20}}c>{\columncolor{gray!20}}c}
\toprule[1.5pt]
& & & \textbf{HydraLoRA} & \multicolumn{6}{c}{\textbf{Ours (\model~)}} \\
\multirow{-2}{*}{\textbf{Task}} & \multirow{-2}{*}{\textbf{Base}} & \multirow{-2}{*}{\textbf{LoRA}} & L=32 & L=28 & L=24 & L=20 & L=16 & L=12 & L=8 \\
\midrule
Boolean Expressions & 73.60 & 74.40 & 75.20 & 77.20 & 76.01 & 76.40 & \textbf{77.60} & 74.07 & 71.35 \\
Causal Judgement & 47.06 & 54.90 & 57.22 & 56.15 & 50.72 & 57.14 & 50.27 & 55.08 & \textbf{59.89} \\
Date Understanding & 37.60 & 40.00 & 44.80 & 55.60 & 54.77 & 56.40 & 57.20 & 59.60 & \textbf{60.80} \\
Disambiguation QA & 34.80 & 45.20 & 56.80 & 60.00 & 65.60 & 64.00 & 64.40 & 53.17 & \textbf{66.00} \\ 
Dyck Languages  & 10.80 & 13.20 & 14.40 & \textbf{15.60} & 13.20 & 14.06 & 14.92 & 14.49 & 14.00 \\
Formal Fallacies  & 44.80 & 46.00 & 46.80 & 46.40 & 46.80 & 47.60 & 48.00 & 47.15 & \textbf{49.20} \\
Geometric Shapes  & 9.70 & 10.20 & 15.20 & 14.40 & \textbf{25.60} & 19.60 & 12.40 & 12.77 & 12.00 \\
Hyperbaton  & 30.80 & 40.00 & 48.40 & 48.80 & 48.40 & 52.00 & \textbf{53.20} & 48.40 & 49.20 \\
Logical Deduction (five objects)  & 22.80 & 33.30 & 45.60 & \textbf{46.80} & 45.13 & 45.53 & 46.34 & 43.10 & 46.00 \\
Logical Deduction (seven objects) & 16.00 & 22.40 & 32.00 & 31.60 & 30.80 & \textbf{32.40} & 30.40 & 30.00 & 29.60 \\
Logical Deduction (three objects) & 35.20 & 41.40 & 44.40 & 68.40 & 59.60 & 61.20 & 52.00 & 56.40 & \textbf{69.60} \\
Movie Recommendation  & 53.50 & 63.05 & 68.67 & \textbf{69.88} & 66.51 & 69.08 & 68.27 & 65.72 & 65.72 \\
Multistep Arithmetic  & 0.80 & 0.80 & 1.20 & \textbf{1.60} & 1.36 & 1.26 & 1.26 & 1.26 & 1.20 \\
Navigate & 42.40 & 52.70 & 57.10 & 54.40 & 54.40 & 62.80 & 62.80 & 58.40 & \textbf{65.20} \\
Object Counting & 40.10 & 44.00 & 42.40 & 44.00 & 45.20 & 46.00 & 44.80 & 46.00 & \textbf{50.00} \\
Penguins in a Table & 21.70 & 22.60 & 26.03 & 28.08 & 33.56 & 28.08 & 28.77 & 33.56 & \textbf{48.63} \\
Reasoning about Colored Objects & 19.40 & 27.20 & 35.60 & 40.00 & 36.59 & 42.00 & 41.60 & \textbf{42.40} & 37.60\\
Ruin Names & 25.40 & 28.70 & 30.65 & 33.47 & 31.92 & \textbf{37.90} & 34.68 & 34.27 & 29.03 \\
Salient Translation Error Detection & 11.20 & 25.20 & 26.80 & 26.00 & 27.95 & \textbf{32.40} & 32.00 & 32.00 & 30.40 \\
Snarks & 44.00 & 44.00 & 46.63 & \textbf{48.31} & 46.63 & 46.63 & 47.19 & 46.63 & 46.63 \\
Sports Understanding & 50.00 & 57.20 & 65.60 & \textbf{67.20} & 58.00 & 57.60 & 54.80 & 55.20 & 58.00 \\
Temporal Sequences & 21.10 & 32.60 & 33.20 & 33.60 & 34.80 & 32.90 & 32.90 & 33.60 & \textbf{36.40} \\
Tracking Shuffled Objects (five objects) & 21.90 & 31.20 & 37.20 & 37.60 & 38.40 & \textbf{40.00} & 38.80 & 38.40 & 39.20 \\
Tracking Shuffled Objects (seven objects) & 14.80 & 14.00 & 26.00 & 27.60 & \textbf{29.60} & 26.40 & 28.80 & 28.40 & 27.60 \\
Tracking Shuffled Objects (three objects) & 41.20 & 51.20 & 71.20 & 71.60 & \textbf{72.80} & 70.00 & 69.20 & 67.60 & 65.60 \\
Web of Lies & 48.80 & 51.20 & 50.80 & 50.00 & 51.97 & 52.80 & 51.20 & \textbf{53.20} & 50.40 \\
Word Sorting & 22.00 & 21.20 & 22.40 & 20.40 & 22.00 & 22.40 & \textbf{24.40} & 23.60 & 22.00 \\
\midrule[0.75pt]
 Avg Performance (EM) & 31.17 & 36.59 & 41.57 & 43.51 & 42.73 & 44.24 & 43.27 & 42.76 & \textbf{44.49} \\
\# of A/B for training/inference & 0/0 & 1/1 & 1/6 & 1/6 & 1/6 & 1/6 & 1/6 & 1/6 & 1/6\\
\% Params & - & 0.062 & 0.205 & 0.179 & 0.153 & 0.128 & 0.102 & 0.077 & 0.051 \\
\bottomrule[1.5pt]
\end{tabular}}
\label{app:bbh_results}
\end{table*}

\begin{figure}[h]
    \centering
    \includegraphics[width=0.9\linewidth]{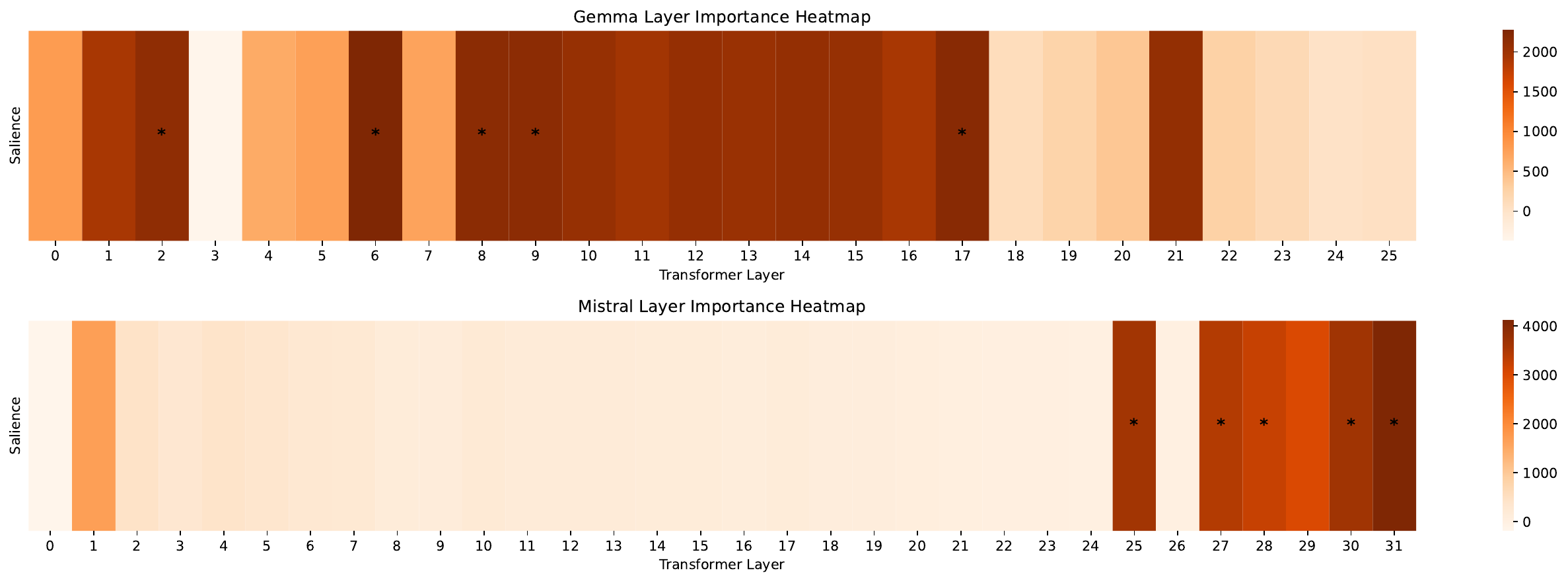}
    \vspace{-0.5em}
    \caption{\small Layer importance heatmaps for Gemma2-2B (top) and Mistral-7B (bottom), highlighting critical adaptation layers (8-17 for Gemma2-2B; 27-31 for Mistral-7B). Saliency values reflect the contribution of each layer, with darker hues indicating higher importance.
    }
    \label{fig:gemma_mistral_imp}
    \vspace{-1em}
\end{figure}

\section{Broader Impacts}\label{app:impact}
\noindent{\textbf{Positive Societal Impacts.}}
The proposed \model~ framework, with its Fisher information-aware layer selection and Bayesian optimization-driven dynamic rank allocation, advances the accessibility and efficiency of LLMs. By optimizing layer-specific adaptation and minimizing parameter growth, \model~ reduces computational barriers, enabling researchers and organizations with limited resources to deploy LLMs effectively in specialized domains. This could accelerate innovation in critical fields such as healthcare, education, and climate science, where domain-specific adaptation is essential. Furthermore, \model~’s parameter-efficient design significantly lowers energy consumption during fine-tuning, aligning with sustainable AI practices and reducing the environmental footprint of model development. Finally, by isolating task-specific features and mitigating domain interference, \model~ enhances the reliability of LLMs in multi-task scenarios, paving the way for safer and more trustworthy AI systems.

\noindent{\textbf{Negative Societal Impacts.}}
While \model~ improves the efficiency of LLM adaptation, its broader deployment may amplify existing risks associated with AI technologies.
By lowering the barrier for domain-specific fine-tuning, it empowers not only beneficial applications but also the rapid creation of sophisticated misinformation by malicious actors.
Although \model~ itself does not directly introduce these issues, its efficiency in adapting LLMs underscores the urgent need for careful scrutiny and responsible use.

\end{document}